\documentclass{article}

\usepackage[preprint,nonatbib]{neurips_2022}

\usepackage[utf8]{inputenc} 
\usepackage[T1]{fontenc}    
\usepackage{hyperref}       
\usepackage{url}            
\usepackage{booktabs}       
\usepackage{amsfonts}       
\usepackage{nicefrac}       
\usepackage{microtype}      
\usepackage{xcolor}         

\usepackage{graphicx}
\usepackage{amsmath}
\usepackage{bm}
\usepackage{multirow}
\usepackage{algorithm}
\usepackage{algorithmic}

\usepackage{subcaption}
\usepackage{wrapfig}

\newcommand{\etal}{\textit{et al}. }
\newcommand{\eg}{\textit{e}.\textit{g}.}

\title{Multi-View Correlation Consistency for Semi-Supervised Semantic Segmentation}

\author{%
  Yunzhong Hou \quad Stephen Gould \quad Liang Zheng \\
  Australian National University\\
  \texttt{\{firstname.lastname\}@anu.edu.au} \\
}

\begin{document}

\maketitle

\begin{abstract}
Semi-supervised semantic segmentation needs \textit{rich} and \textit{robust} supervision on unlabeled data. Consistency learning enforces the same pixel to have similar features in different augmented views, which is a robust signal but neglects relationships with other pixels. In comparison, contrastive learning considers rich pairwise relationships, but it can be a conundrum to assign binary positive-negative supervision signals for pixel pairs. In this paper, we take the best of both worlds and propose multi-view correlation consistency (MVCC) learning: it considers \textit{rich} pairwise relationships in self-correlation matrices and matches them across views to provide \textit{robust} supervision. Together with this correlation consistency loss, we propose a view-coherent data augmentation strategy that guarantees pixel-pixel correspondence between different views. In a series of semi-supervised settings on two datasets, we report competitive accuracy compared with the state-of-the-art methods. Notably, on Cityscapes, we achieve 76.8\% mIoU with 1/8 labeled data, just 0.6\% shy from the fully supervised oracle.

\end{abstract}

\section{Introduction}
\label{sec:intro}

\begin{figure}
\centering
\begin{subfigure}[t]{0.25\linewidth}
\includegraphics[width=\linewidth]{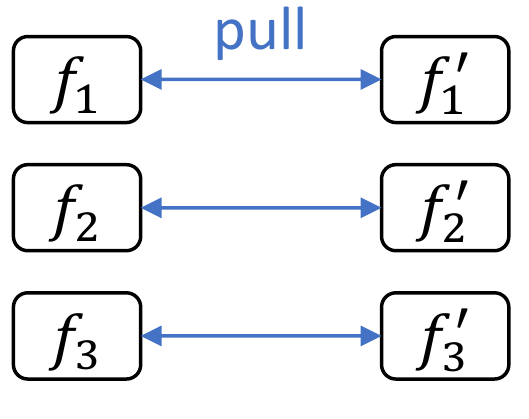}
\caption{Consistency}
\end{subfigure}
\hfill
\begin{subfigure}[t]{0.25\linewidth}
\includegraphics[width=\linewidth]{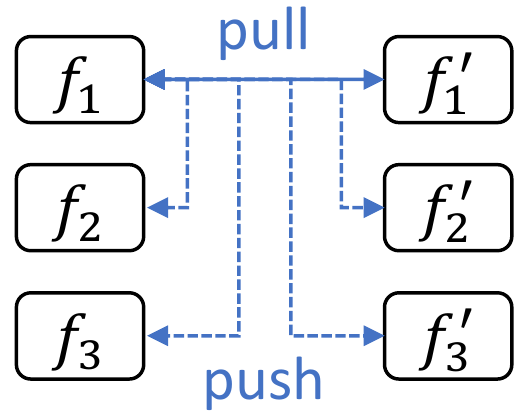}
\caption{Contrastive}
\end{subfigure}
\hfill
\begin{subfigure}[t]{0.25\linewidth}
\includegraphics[width=\linewidth]{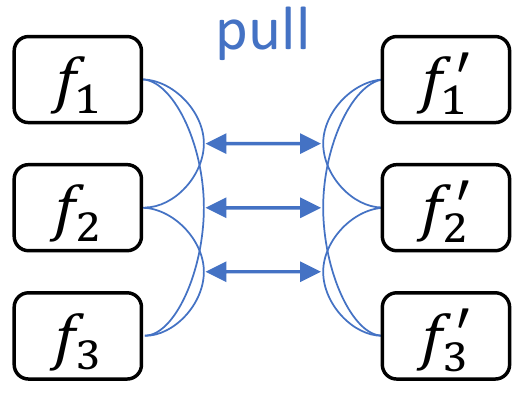}
\caption{Correlation-consistency}
\end{subfigure}
\caption{
An illustration of three ways to utilize unlabeled data for supervision. 
Given features of three data points ($n\in\left\{1,2,3\right\}$) in two augmented views ($\left\{\bm{f}_n\right\}$ and $\left\{\bm{f}'_n\right\}$), 
{(a)} consistency learning provides \textit{robust} supervision by maintaining similar features of the {same} data point across views, but disregards relationships between {different} data points. 
{(b)} On the other hand, contrastive learning considers \textit{rich} pairwise relationships by repelling features of different data points, but lacks a robust way to decide positive or negative assignment in semi-supervised semantic segmentation. 
{(c)} The 
proposed correlation consistency learning uses pixel pairs \emph{richly} and 
\textit{robustly} supervises the self-correlation matrices to be similar across views. 
}
\label{fig:demo}
\end{figure}

In semi-supervised learning, the most important challenge is how to effectively utilize unlabeled data. Consistency~\cite{berthelot2019mixmatch,sohn2020fixmatch,tarvainen2017mean} and contrastive learning~\cite{he2020momentum,chen2020simple} are two popular strategies. 
For image classification, given augmented views of unlabeled images (Fig.~\ref{fig:demo}), consistency learning enforces different views of the same image to have similar features, whereas 
contrastive learning encourages the network output of a certain image to be dissimilar to all but those of the same image. 


Semi-supervised semantic segmentation, the problem studied in this paper, requires \textit{rich} and \textit{robust} supervision on unlabeled data. 
By minimizing feature distance for pixels from the same location in different views, 
consistency learning~\cite{french2019semi,ouali2020semi,mendel2020semi} establishes very \textit{reliable} supervision signals. 
However, this approach does not provide sufficiently abundant supervision, as it ignores relationships with pixels in different locations. 
Contrastive learning~\cite{zhao2021contrastive,zhou2021c3,zhong2021pixel,liu2022reco,alonso2021semi}, on the other hand, employs \textit{ample} pairwise supervision by promoting (punishing) feature similarities of {positive} ({negative}) pixel pairs. 
Nonetheless, deciding binary positive-negative assignments can be non-trivial.
First, if we treat each pixel as a distinct class~\cite{wang2021dense,zhao2021contrastive} as done in image classification, the neighboring pixels would be forced to have dissimilar features. This is clearly undesirable because these pixels usually describe the same object and are strongly correlated. 
Second, some works use pseudo labels to decide pixel semantics and in turn assign binary {positive-negative} pairs~\cite{zhou2021c3,zhong2021pixel,alonso2021semi,liu2022reco}, but this process is often hindered by the noisy nature of pseudo labels. Small errors in pseudo labels may reverse the binary positive-negative pair assignment, completely overturning feature similarity supervision. 
To obtain \textit{rich} and \textit{robust} supervision, 
we propose multi-view correlation consistency (MVCC) learning, which achieves state-of-the-art results in multiple settings. In a nutshell, we introduce a correlation consistency loss to enforce self-correlation matrices to be similar between views  (Fig.~\ref{fig:demo}). Compared to consistency learning, our method computes similarities between a much greater number of pixel pairs and thus benefits from a \textit{richer} description of the data distribution. Compared with contrastive learning, importantly, pixel-pixel similarity no longer increases (decreases) according to the positive (negative) assignment; 
instead, it is supervised by the similarity of the same pixel pair in another augmented view. Apparently, the proposed supervision is relatively weak, but the way it is formulated allows for \textit{robust} pairwise supervision, avoiding the each-pixel-as-a-separate-class assumption and mitigating the influence of noisy pseudo labels. 
Apart from the loss function, MVCC introduces a view-coherent augmentation strategy that provides pixel-pixel correspondence between views for \emph{image-level}~\cite{xie2021propagate,lai2021semi} and \textit{region-level}~\cite{french2019semi,olsson2021classmix,hu2021semi} augmentation.
For the former, we use sampling grids of affine transformations and implement invertible and differentiable geometric augmentation, \eg, random cropping, scaling, and flipping, with affordable computation cost. For the latter, we restrict region-level augmentation methods (\textit{e.g.}, CutMix \cite{yun2019cutmix}) to be view coherent, so that for each area, we can find its correspondence in other views. 
In addition, this augmentation strategy is complementary to \textit{pixel-level} augmentation \cite{ke2020guided,ouali2020semi}. 

 



\section{Related work}
\label{sec:related_work}


\textbf{Consistency and contrastive learning in semi-supervised semantic segmentation.} 
\textit{Consistency} learning usually follows the same pipeline in classification: either directly minimizes feature distances~\cite{tarvainen2017mean,miyato2018virtual,french2019semi,ouali2020semi}, or uses
a common target as middle ground~\cite{laine2016temporal,berthelot2019mixmatch,berthelot2019remixmatch}. 
On the other hand, for \textit{contrastive} learning, some follow the image classification pipeline and treat each pixel as a distinct class \cite{wang2021dense,zhao2021contrastive}. Others believe neighboring pixels are correlated and should not be forced to have distinct features and thus use pixel classes to construct \textit{positive} and \textit{negative} pairs~\cite{wang2021exploring,zhang2021looking,hu2021region}. In the absence of human labels, pseudo labels are considered for this purpose~\cite{zhou2021c3,zhong2021pixel,alonso2021semi}, but their unreliable quality would potentially limit system performance.
In this work, the proposed correlation consistency loss does not treat every pixel as a class or use pseudo labels to give binary pairs. 

\textbf{Data augmentation} for segmentation can be performed on the pixel, region, and image levels. 
Pixel-level augmentation intends to add noise on images or feature maps \cite{ke2020guided,ouali2020semi}.
Region-level augmentation usually replaces parts of an image with another image, such as MixUp \cite{zhang2018mixup} (overlay an image onto another), CutMix \cite{yun2019cutmix,french2019semi} (replace image region),  ClassMix \cite{olsson2021classmix} (CutMix based on pseudo labels), and AEL \cite{hu2021semi} (class-balanced extension to Copy-Paste \cite{ghiasi2021simple} and CutMix). 
Image-level augmentation uses geometric transformations, \eg, cropping, scaling, and flipping.  
For the three strategies, a common requirement is the pixel-pixel correspondence between augmented views, but it \textbf{a)} is costly to compute pixel-pixel distances in image-level augmentation  \cite{xie2021propagate,lai2021semi}, and \textbf{b)} is often not satisfied in region-level augmentation. We solve these problems by introducing efficient 
{image-level} augmentation and view-coherent extension of {region-level} augmentation.

\textbf{Pseudo label} is a popular choice for entropy minimization in semi-supervised learning, 
which can be either hard (one-hot) labels \cite{lee2013pseudo,han2018coteaching,sohn2020fixmatch,hung2018adversarial,chen2021semi,he2021re} or soft labels \cite{berthelot2019mixmatch,xie2019unsupervised,chen2020big,yuan2021simple}. It may be obtained from the same network used in training \cite{berthelot2019mixmatch,sohn2020fixmatch}, a temporal ensemble network using exponential moving average (EMA) \cite{tarvainen2017mean,french2019semi}, or a pair of differently initialized networks \cite{han2018coteaching,chen2021semi}. 
Following MixMatch, our system uses the average Softmax probabilities from EMA models as pseudo labels. 


\section{Method}
\label{sec:method}

\subsection{Overview}
\label{secsec:overview}
We adopt the semi-supervised learning pipeline MixMatch \cite{berthelot2019mixmatch}, a consistency learning method that can effectively leverage unlabeled data over multiple augmented views. Note that we do not use the MixUp augmentation~\cite{zhang2018mixup} because it is found less effective in segmentation \cite{islam2020feature}.
Suppose we have a segmentation network $f\left(\cdot\right)$ and two different views $\bm{x}_\text{U}$ and $\bm{x}'_\text{U}$ of an unlabeled image. 
MixMatch first averages the output from the EMA temporal ensemble network $\bar{f}\left(\cdot\right)$ to formulate the soft pseudo label 
$\hat{\bm{y}}_\text{U} = \left(\bar{f}\left(\bm{x}_\text{U}\right)+\bar{f}\left(\bm{x}'_\text{U}\right)\right)/2$.
With this common pseudo label $\hat{\bm{y}}_\text{U}$, two views $\bm{x}_\text{U}$ and $\bm{x}'_\text{U}$ of the unlabeled image, and the human-annotated image-label pair $\left<\bm{x}_\text{L},\bm{y}_\text{L}\right>$, the network $f\left(\cdot\right)$ is optimized with the following loss functions,
\begin{align}
\label{eq:loss_sup}
\mathcal{L}_\text{L} & = H\left(f\left(\bm{x}_\text{L}\right),\bm{y}_\text{L}\right),\\
\label{eq:loss_unsup}
\mathcal{L}_\text{U} & = \lVert f\left(\bm{x}_\text{U}\right)-\hat{\bm{y}}_\text{U}\rVert_2^2 + \lVert f\left(\bm{x}'_\text{U}\right)-\hat{\bm{y}}_\text{U}\rVert_2^2,
\end{align}
where $H\left(\cdot,\cdot\right)$ denotes the cross-entropy loss and $\lVert\cdot\rVert_2$ the $L_2$ norm. As long as $f\left(\bm{x}_\text{U}\right)$ and $f\left(\bm{x}'_\text{U}\right)$ are pixel-pixel corresponded, the consistency regularization in Eq.~\ref{eq:loss_unsup} is robust. 
Based on this pipeline, we introduce a correlation consistency loss and a view-coherent data augmentation strategy, constituting multi-view correlation consistency (MVCC) learning. In the following, we detail the components. 

\subsection{Correlation-consistency loss}
\label{secsec:correlation-consistency}
As mentioned, it is desirable to have \textit{rich} and \textit{robust} supervision in semi-supervised semantic segmentation. 
While the consistency loss  $\mathcal{L}_\text{U}$ provides \textit{robust} supervision, 
it neglects relationships between non-corresponding pixels. 
To obtain \textit{rich} supervision, contrastive learning uses pairwise relationships, and InfoNCE loss \cite{van2018representation} is often adopted. 
Specifically, this loss function directly enforces similar (distinct) features from {positive} (negative) pairs, 
\begin{align}
\label{eq:loss_infonce}
     \mathcal{L}_\text{NCE} = \sum_{\bm{f}}{\sum_{\bm{f}^+}{- \log \frac{\exp\left(\bm{f}\cdot \bm{f}^+/\tau\right)}{\exp\left(\bm{f}\cdot \bm{f}^+/\tau\right) + \sum_{\bm{f}^-}{\exp\left(\bm{f}\cdot \bm{f}^-/\tau\right)}}}},
\end{align}
where $\tau$ denotes the temperature hyper-parameter, and $\bm{f}^+$ and $\bm{f}^-$ form positive and negative pairs, respectively, with a certain feature $\bm{f}$. 
The binary positive-negative pair assignment directly influences feature similarity learning, but is difficult to decide: treating each pixel as a distinct class~\cite{zhao2021contrastive} breaks the correlation between neighboring pixels; using pseudo labels to assign positive and negative pairs~\cite{zhou2021c3,zhong2021pixel,alonso2021semi} is susceptible to their noisy nature. 

\begin{figure}
\begin{minipage}[c]{0.55\textwidth}
\centering
\begin{subfigure}[b]{0.47\linewidth}
\includegraphics[width=\linewidth]{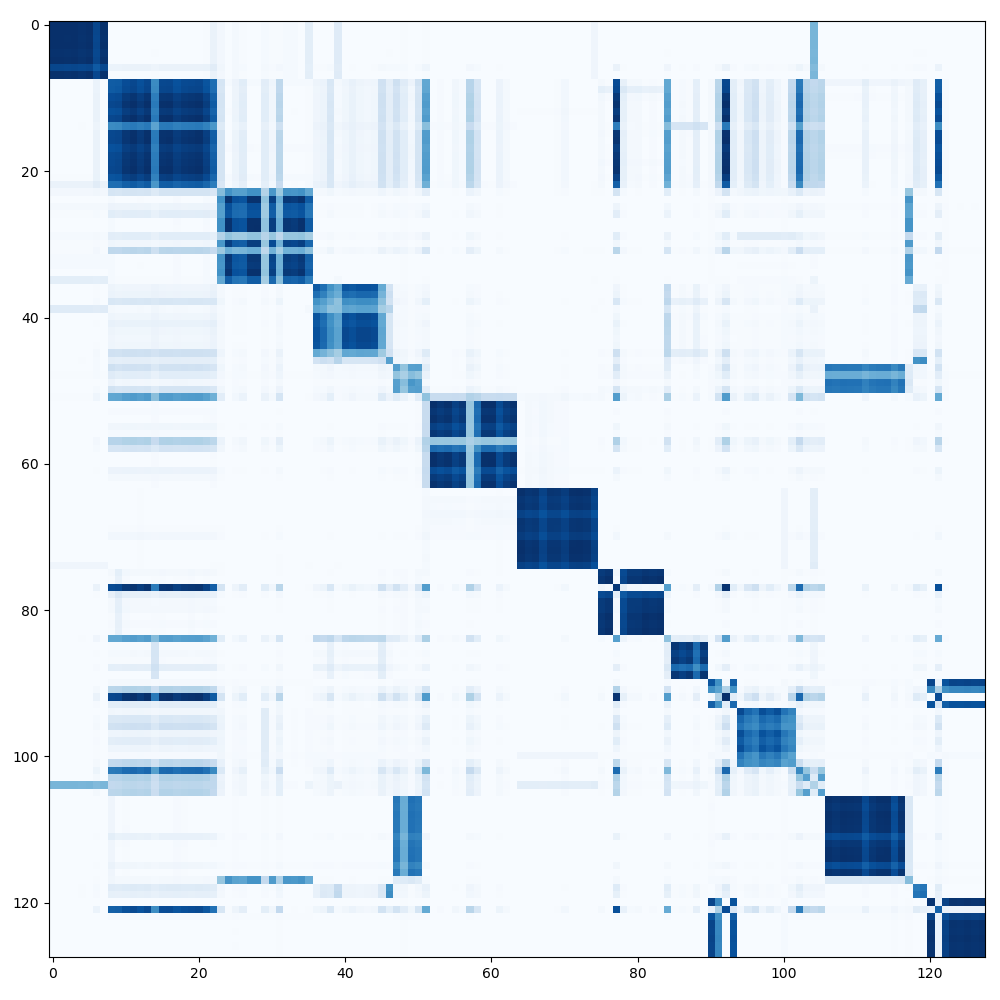}
\end{subfigure}
\hfill
\begin{subfigure}[b]{0.47\linewidth}
\includegraphics[width=\linewidth]{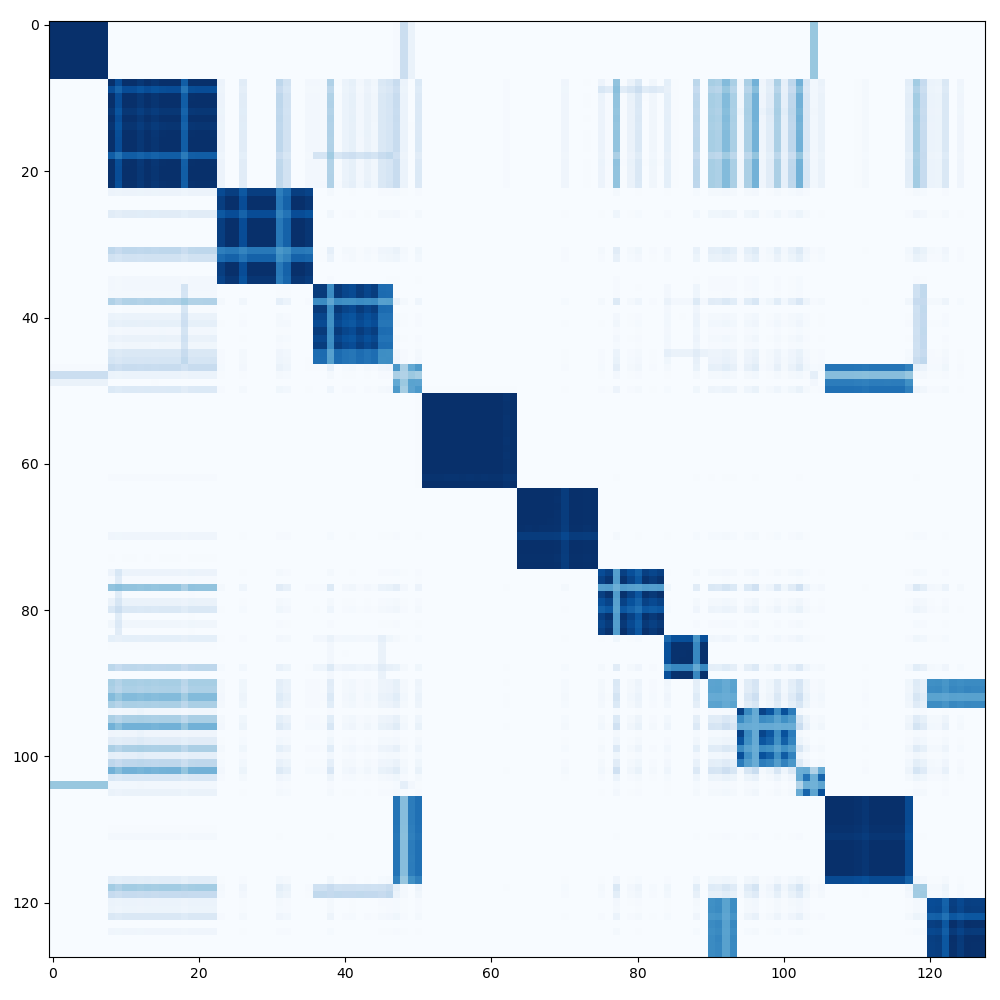}
\end{subfigure}
\end{minipage}
\hfill
\begin{minipage}[c]{0.4\textwidth}
\vspace{3mm}
\caption{Examples of self-correlation matrices $A_\text{feat}$ and $A_\text{ema}'$ used in the correlation consistency loss. Given features and their EMA model targets over two augmented views ($\left\{\bm{f}_n\right\}$ and $\left\{\bar{\bm{f}}'_n\right\}$), we calculate their self-correlation matrices $A_\text{feat}$ and $A_\text{ema}'$ (Eq.~\ref{eq:correlation}), respectively, and then maintain their consistency. A cartoon illustration is  in Fig.~\ref{fig:demo} (c).
}
\label{fig:correlation}
\end{minipage}
\end{figure}

Motivated by the above discussions, we investigate more \textit{robust} supervision for \textit{rich} pairwise relationships. Specifically, we propose the correlation consistency loss, which no longer regularizes feature similarities based on binary positive-negative pair assignment. Instead, it keeps the  feature correlation maps in two views (see Fig.~\ref{fig:correlation}) to be similar.

\subsubsection{Loss function}
\label{secsecsec:loss}

Mathematically, we formulate the correlation consistency loss as follows. 
Given $N$ pixels in a certain view that have $D$-dimensional features $\left\{\bm{f}_n\right\}, \bm{f}_n\in \mathbb{R}^D, n\in\left\{1,...,N\right\}$, first, we choose the EMA model features on the opposite view $\left\{\bar{\bm{f}}'_n\right\}$ as their target, and 
stack them respectively to formulate $\left\{\bm{f}_n\right\}\rightarrow \mathcal{F} \in \mathbb{R}^{N\times D}$ and its target $\left\{\bar{\bm{f}}'_n\right\}\rightarrow \bar{\mathcal{F}}'$. 
Next, we calculate their pairwise relationships,
\begin{align}
\label{eq:correlation}
     A_\text{feat} = \mathcal{F}\cdot\mathcal{F}^T, \;
     A_\text{ema}' = \bar{\mathcal{F}}'\cdot \bar{\mathcal{F}}'^T,
\end{align}
where $A_\text{feat}, A_\text{ema}' \in \mathbb{R}^{N \times N}$ denote self-correlation matrices. Then, similar to the Softmax operation in Eq.~\ref{eq:loss_infonce}, we normalize the similarity between each pixel and all others with the $L_2$ norm, 
\begin{align}
\label{eq:l2norm}
     \widetilde{A}_{\text{feat}\left[i,:\right]} = \frac{A_{\text{feat}\left[i,:\right]}}{\left\|A_{\text{feat}\left[i,:\right]} \right\|_2}, \;
     \widetilde{A}'_{\text{ema}\left[i,:\right]} = \frac{A'_{\text{ema}\left[i,:\right]}}{\left\|A'_{\text{ema}\left[i,:\right]} \right\|_2},
\end{align}
where $\left[i,:\right]$ denotes all the elements in the $i$-th row. 
Lastly, we repeat Eq.~\ref{eq:correlation} and Eq.~\ref{eq:l2norm} for the other view and compute the correlation consistency loss between self-correlation matrices,
\begin{align}
\label{eq:loss_cc}
     \mathcal{L}_\text{CC} = \frac{1}{N} \left( \left\|\widetilde{A}_\text{feat} - \widetilde{A}_\text{ema}' \right\|_F^2 +  \left\|\widetilde{A}'_\text{feat} - \widetilde{A}_\text{ema} \right\|_F^2 \right),
\end{align}
where $\left\|\cdot\right\|_F$ denotes the Frobenius norm (entry-wise $L_2$ norm for matrix). 

To balance the loss terms (Eq~\ref{eq:loss_sup}, Eq.~\ref{eq:loss_unsup}, and Eq.~\ref{eq:loss_cc}), we set the final loss as 
\begin{align}
\label{eq:loss_overall}
     \mathcal{L} = \mathcal{L}_\text{L}+0.1\times\mathcal{L}_\text{U}+0.1\times \mathcal{L}_\text{CC}.
\end{align}

\subsubsection{Category-normalized data sampling}
\label{secsecsec:sampling}

Two issues exist with the proposed correlation consistency loss. \textbf{a)} The complexity for computing pairwise relationships $\mathcal{O}\left(N^2\right)$ scales poorly with the number of pixels $N$. \textbf{b)} There is a long-tailed pixel distribution over classes \cite{he2021re,hu2021semi}. 

To tackle them both, we introduce category-normalized data sampling, which chooses a similar number of pixels for each class. 
First, for each mini-batch of unlabeled images, we compute the empirical categorical distribution over $C$ classes as $\left\{p_c\right\}, c\in\left\{1,...,C\right\}$, based on their pseudo labels. 
Then, each pixel $j$ is given a sampling probability $q_j\propto 1/p_c$ that is proportional to the inverse of the empirical probability for its class $c$ (using its pseudo label). Finally, for Eq.~\ref{eq:correlation}, we randomly choose $N$ pixels over the mini-batch according to the sampling probability $q_j$ of each pixel $j$.  
Using this trick, we can efficiently sample a manageable number of pixels to avoid computation cost explosion, and at the same time make sure that the sampled pixels have a relatively balanced class distribution. 

\subsubsection{Features for correlation matrices}
\label{secsecsec:choice}

To obtain the $D$-dimensional features $\left\{\bm{f}_n\right\}$, we sample feature vectors from the $C$-dimensional Softmax output maps $f\left(\cdot\right)$ of both views, so $D=C$. In our preliminary experiment, we do not witness a significant accuracy difference when using features from additional projection heads.

For their consistency targets  $\left\{\bar{\bm{f}}'_n\right\}$, instead of EMA network output from the opposite view, we use the pseudo labels $\left\{\left(\bar{\bm{f}}_n+\bar{\bm{f}}'_n\right)/2\right\}$ due to two considerations. 
\textbf{a)} The soft pseudo label contains information from both views and has better stability.
\textbf{b)} It offers a common target for one pixel across two views, which provides a middle ground to help enforce consistency \cite{berthelot2019mixmatch,berthelot2019remixmatch}. 
In this setting, although $\mathcal{L}_\text{CC}$ also uses pseudo labels as supervision signals like $\mathcal{L}_\text{NCE}$, its consistency-based implementation makes it more robust to label noise. See Section~\ref{secsec:discussion} for more discussions. 


\begin{figure}
\centering
\begin{subfigure}[t]{0.22\linewidth}
\includegraphics[width=\linewidth]{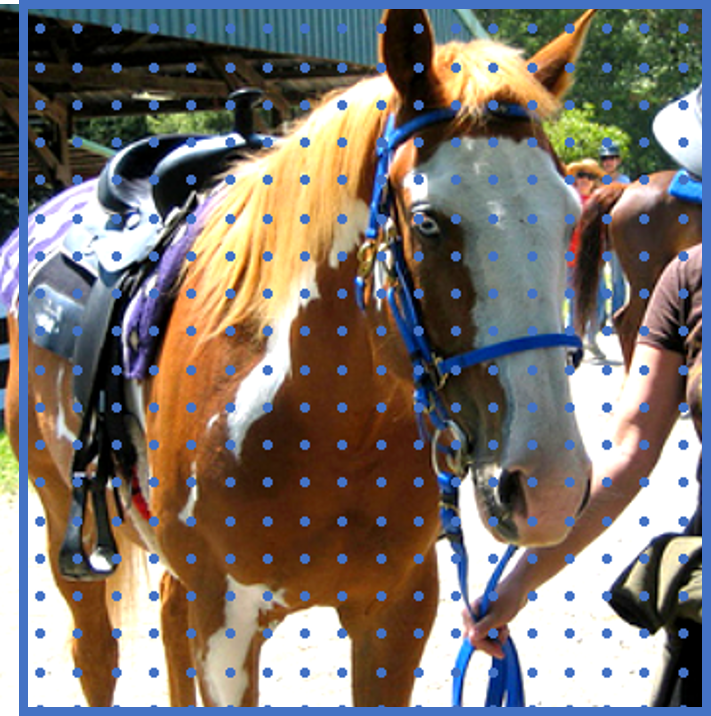}
\caption{}
\end{subfigure}
\hfill
\begin{subfigure}[t]{0.22\linewidth}
\includegraphics[width=\linewidth]{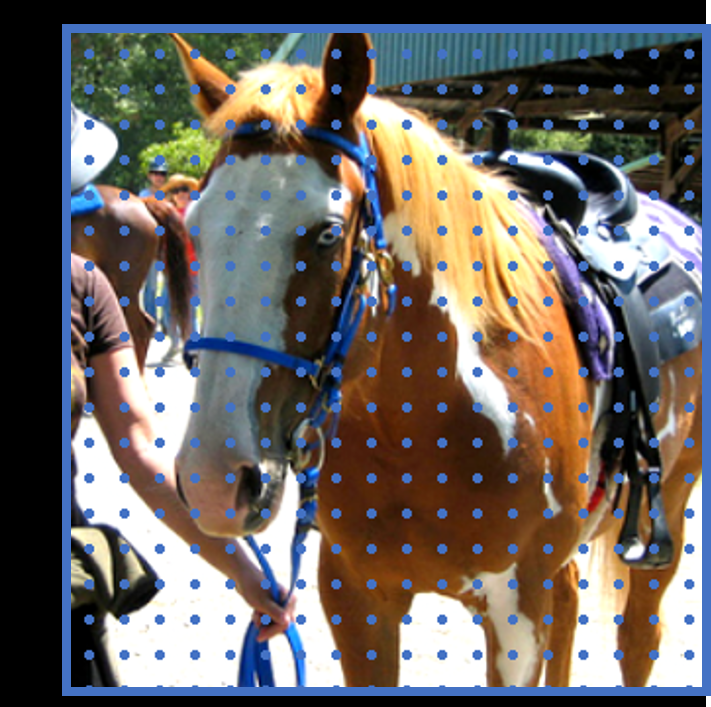}
\caption{}
\end{subfigure}
\hfill
\begin{subfigure}[t]{0.22\linewidth}
\includegraphics[width=\linewidth]{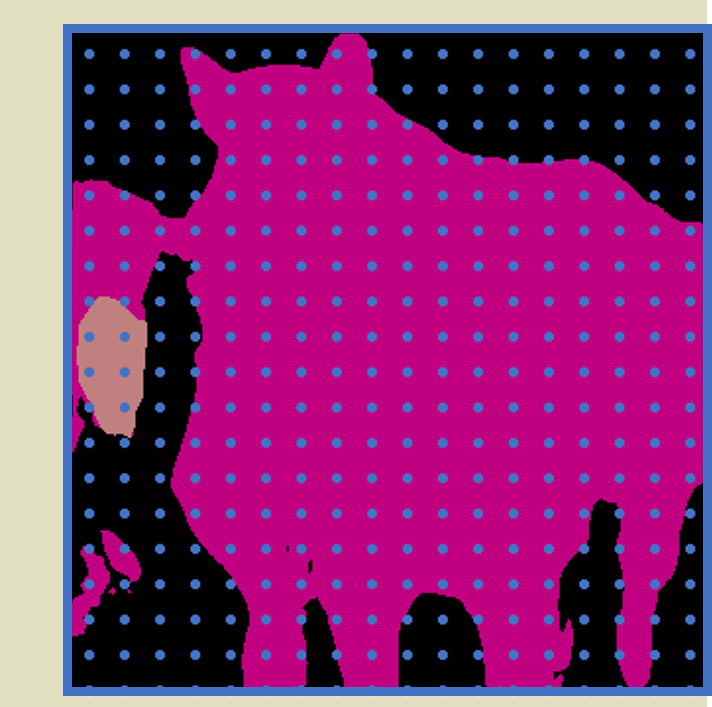}
\caption{}
\end{subfigure}
\hfill
\begin{subfigure}[t]{0.22\linewidth}
\includegraphics[width=\linewidth]{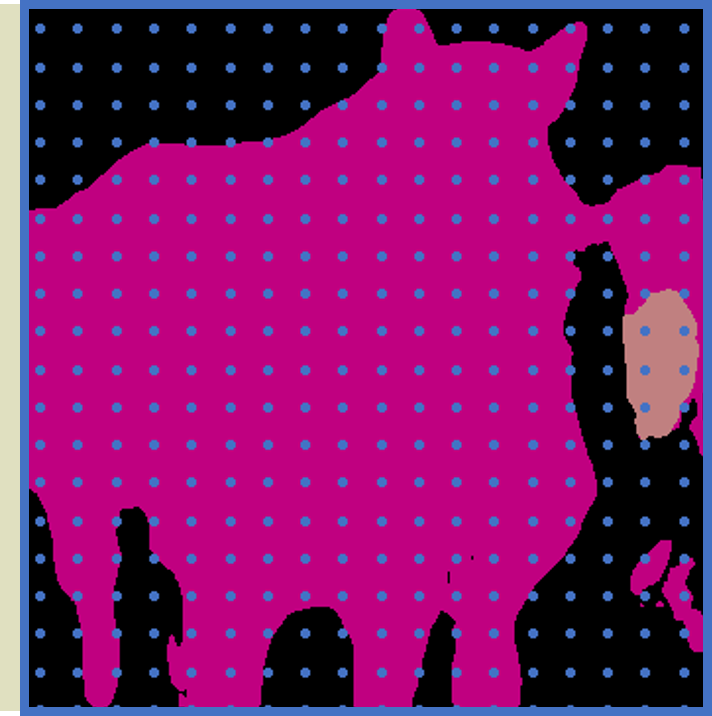}
\caption{}
\end{subfigure}
\caption{An illustration of the proposed image-level augmentation method. We augment an image {(a)} to image {(b)} through affine transformation, which is implemented by sampling grids (blue dots). From {(b)}, we extract its feature map {(c)}. To restore pixel-pixel correspondence between {(a)} and {(d)}, we apply inverse affine transformation (through another sampling grid) to obtain {(d)} from {(c)}. 
}
\label{fig:sampling_grid}
\end{figure}

\subsection{View-coherent data augmentation}
\label{secsec:augmentation}

\subsubsection{Image-level data augmentation}
\label{secsecsec:image-level}
{Image-level} geometric augmentation methods like cropping, flipping, and scaling 
are found effective for semantic segmentation, but seldom applied in semi-supervised settings due to the high cost of computing pixel-pixel correspondence. 
In fact, existing methods \cite{xie2021propagate,lai2021semi} go through every pixel in every view to calculate their distances in a common coordinate system, and then decide correspondence based on a certain threshold, which has $\mathcal{O}\left(N^2\right)$ computation complexity. 

To reduce complexity in obtaining pixel-pixel correspondence, 
we use sampling grids \cite{jaderberg2015spatial} to implement affine transformations, which are differentiable and invertible. 
As shown in Fig.~\ref{fig:sampling_grid}, for pixels in one view, 
sampling grids compute their coordinates in another view via affine transformations. 
After extracting feature map of the augmented image, we invert the applied augmentation via another sampling grid, returning a feature map that is pixel-pixel corresponded with the original image. From Fig.~\ref{fig:sampling_grid} (a) to Fig.~\ref{fig:sampling_grid} (b) and from Fig.~\ref{fig:sampling_grid} (c) to Fig.~\ref{fig:sampling_grid} (d), the computation complexity is both $\mathcal{O}\left(N\right)$, so the overall complexity is $\mathcal{O}\left(N\right)$. 
In practice, given two views, we use this implementation for image-level augmentation to produce two sets of segmentation maps that are pixel-pixel corresponded (Fig.~\ref{fig:system}). Note it is from such feature maps that we generate the pseudo labels $\hat{\bm{y}}_\text{U}$. 


\subsubsection{Region-level data augmentation}
\label{secsecsec:region-level}

{Region-level} augmentation modifies an area of an image. 
For views generated by different {image-level} geometric augmentation, if applied randomly, region-level augmentation easily breaks coherency between views, leading to non-corresponded pixels. 

\begin{figure}
\centering
\includegraphics[width=\linewidth]{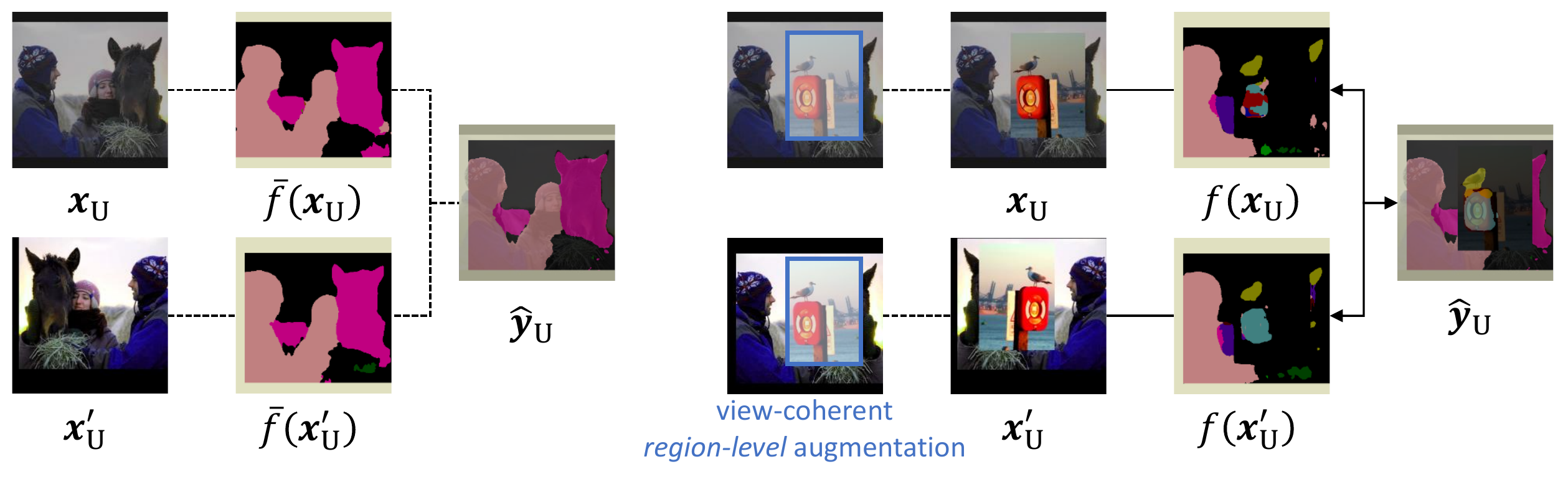}
\caption{Flow of unlabeled images in multi-view correlation consistency (MVCC) learning. 
\textbf{Left}: we feed two views to the EMA model 
and then average the pixel-pixel corresponded feature maps $\bar{f}\left(\bm{x}_\text{U}\right)$ and $\bar{f}\left(\bm{x}'_\text{U}\right)$ to obtain pseudo label $\hat{\bm{y}}_\text{U}$. 
\textbf{Right}: for view-coherent region-level augmentation, we first invert \textit{image-level} augmentation, then apply the same \textit{region-level} augmentation, and finally restore the \textit{image-level} augmentation. With pixel-pixel correspondence in $f\left(\bm{x}_\text{U}\right)$ and $f\left(\bm{x}'_\text{U}\right)$, 
we calculate the unsupervised loss (Eq.~\ref{eq:loss_unsup}) and the correlation consistency loss (Eq.~\ref{eq:loss_cc}).
Dotted lines: forward pass only; solid lines: both forward and backward passes; arrows: loss computation. 
}
\label{fig:system}
\end{figure}

To solve this problem, we introduce a view-coherent extension of existing region-level data augmentation. 
Essentially, as shown in Fig.~\ref{fig:system} (right), we first invert the geometric augmentation to provide correspondence. Then, we apply the same region-level augmentation to the two views to maintain correspondence. Lastly, we add the image-level augmentation back to the two views so that the network can still enjoy geometric augmentations. 
The proposed view-coherent extension supports region-level augmentation methods such as CutMix \cite{yun2019cutmix,french2019semi}, ClassMix \cite{olsson2021classmix} and AEL \cite{hu2021semi}. In this work, we use the most basic one, CutMix, for demonstration. 


\subsection{Discussions}
\label{secsec:discussion}


\textbf{Contrastive learning ($\mathcal{L}_\text{NCE}$) is more susceptible to noisy pseudo labels than our method ($\mathcal{L}_\text{CC}$).} 
Both methods promote or punish feature similarities with pseudo labels as supervision signals (Eq.~\ref{eq:loss_infonce} and Section~\ref{secsecsec:choice}), but have different sensitivities to pseudo label noise. Suppose we have a pair of pixels that are incorrectly assigned with slightly different pseudo labels. 
The former might believe they are of different classes and strongly punish the feature similarity due to its binary positive-negative assignment. 
In comparison, the latter applies much softer supervision: maintaining self-correlation between features to be consistent with that of the pseudo labels, and thus reduces the influence of pseudo label noise. 
In fact, the unreliable binary assignment is an innate problem with $\mathcal{L}_\text{NCE}$ in semi-supervised semantic segmentation, which is verified by the wide usage of hard negative sampling in previous contrastive methods~\cite{wang2021exploring,lai2021semi,alonso2021semi,zhou2021c3}. 


\textbf{Scenarios where contrastive learning is most affected.} As mentioned above, contrastive learning is susceptible to pseudo label noise. So it suffers most when pseudo label noise is at a high level. This happens when the labeled data ratio is low and models are too weak to predict accurate pseudo labels.

\textbf{MVCC vs. SwAV \cite{caron2020unsupervised}: no need for clustering.} While both methods rely on consistency, their mechanisms are significantly different. 
SwAV clusters the features and enforces consistency between cluster assignments (correlation between each feature and cluster centers) to avoid the large batch size requirement. 
In comparison, correlation consistency does not need to cluster the data and instead maintains consistency of the self-correlation matrices, which requires less computation. 
In addition, semantic segmentation does not have the large batch size requirement that SwAV deals with: different from images, even with a small batch size, pixels still provide abundant pairwise relationships. 



\textbf{Limitation.} 
Despite its robustness under limited supervision (see Section \ref{secsec:ablation}), $\mathcal{L}_\text{CC}$ applies weaker supervision than $\mathcal{L}_\text{NCE}$. It means for fully-labeled scenarios with reliable human labels, the traditional InfoNCE loss will likely outperform the proposed loss function. 

\textbf{Potential negative social impacts.} 
Semi-supervised learning might allow for model training at a larger scale, which can have a negative environmental impact. 


\section{Experiment}
\label{sec:experiments}

\subsection{Experimental settings}
\label{secsec:exp_setting}

\textbf{Dataset.} 
We verify the efficacy of the proposed MVCC algorithm on the following datasets. 
\textit{Cityscapes} \cite{cordts2016cityscapes} is a dataset for urban street scene understanding. It consists of 2,975 training images and 500 validation images of high resolution among 19 different semantic classes. 
\textit{Pascal VOC 2012} \cite{everingham2010pascal} is a generic object dataset across 21 distinct semantic classes. It contains 1,464 training images and 1,449 validation images of medium resolution. Following common practice, we additionally include the augmented partition \cite{hariharan2011semantic} for a combined training set (10,582 images in total). 
For both datasets, we assume 1/32, 1/16, 1/8, and 1/4 of all the training data are labeled. 

\textbf{Network architecture.} We use DeepLabV3+ \cite{chen2018encoder} with depth-wise separable convolutions. Specifically, it has a ResNet-50 \cite{he2016deep} feature extractor and an output stride of 16. 

\begin{table}[]
\caption{Comparison with state of the arts on Cityscapes. Under each labeled data ratio, we show  mIoU (\%) of the overall system (first column) and improvement (in absolute value) over supervised-only baseline (second column).
Methods with $^*$ are re-implemented by USRN \cite{guan2022unbiased}.  
Highest numbers in each column are highlighted in \textbf{bold}. For our method, standard deviation of three runs is shown. }
\label{tab:city}
\centering
\setlength{\tabcolsep}{3pt}
\small
\begin{tabular}{l|cc|cc|cc|cc|c}
\toprule
Method                          & \multicolumn{2}{c|}{1/32} & \multicolumn{2}{c|}{1/16} & \multicolumn{2}{c|}{1/8} & \multicolumn{2}{c|}{1/4} & Oracle \\ \hline
ECS \cite{mendel2020semi}                       & -           & -          & -           & -          & 67.4       & +4.3        & 70.7       & +3.4        & 74.8   \\
PC2Seg \cite{zhong2021pixel}                   & -           & -          & -           & -          & 72.1       & +4.0        & 73.8       & +0.9        & 73.6   \\
Alonso \etal \cite{alonso2021semi}  & -           & -          & -           & -          & 70.0       & -          & 71.6       & -          & 74.2   \\
ELN \cite{kwon2022semi}                      & -           & -          & -           & -          & 70.3      & \textbf{+10.5}      & 73.5      & \textbf{+11.7}      & 77.7   \\
ST++ \cite{yang2021st++}                     & -           & -          & -           & -          & 72.7       & +6.9        & 73.8       & +5.4        & -      \\
PSMT \cite{liu2021perturbed}                     & -           & -          & -           & -          & 74.4      & +5.5        & 75.2      & +3.4        & -      \\
DBSN$^*$ \cite{yuan2021simple}                      & 62.2        & +2.4        & 67.3        & +3.0        & 73.5       & +4.6        & -          & -          & \textbf{78.3}   \\
CAC \cite{lai2021semi}                      & 62.2        & +2.4        & 69.4        & +5.1        & 69.7       & +3.7        & 72.7       & +2.0        & 77.7   \\
CPS$^*$ \cite{chen2021semi}           & 62.5        & +2.7        & 69.8        & +5.5        & 74.4       & +5.5        & -          & -          & \textbf{78.3}   \\
USRN \cite{guan2022unbiased}                     & 64.6        & +4.8        & 71.2        & +6.9        & 75.0       & +6.1        & -          & -          & \textbf{78.3}   \\ \hline
MVCC (Ours)                & \textbf{72.0} \footnotesize{$\pm$ 1.1}        & \textbf{+16.7}       & \textbf{74.8} \footnotesize{$\pm$ 0.4}        & \textbf{+13.2}      & \textbf{76.8} \footnotesize{$\pm$ 0.1}      & +8.9       & \textbf{77.2}  \footnotesize{$\pm$ 0.1}      & +4.8        & 77.4 \footnotesize{$\pm$ 0.1} \\
\bottomrule
\end{tabular}
\end{table}


\textbf{Implementation details.} For data augmentation, we use random scaling, flipping, cropping ($512\times1024$ for Cityscapes and $360\times360$ for Pascal VOC), and color jitter to produce two different views, and apply CutMix in a view-coherent manner. 
For multiple views, their scaling and translation (cropping offset) differences are maintained within $\left[0.9,1.1\right]$ and $\left[0,0.1\right]$ times the image size, respectively. 
The EMA model update ratio is set as $0.99$ following existing works \cite{french2019semi,laine2016temporal,tarvainen2017mean}. 
In addition, we apply label smoothing with a factor of $0.1$ to stabilize the training following image classification \cite{szegedy2016rethinking,muller2019does,lukasik2020does}.
For category-normalized data sampling, we choose $N=2048$ pixels for both datasets. 

For labeled data and two views of unlabeled data, we use batch sizes of 4 and 8 for Cityscapes and Pascal VOC, respectively. Learning rates are set to $0.05$ and $0.01$ for the two datasets, and the feature extractor has $0.1\times$ the base learning rates. We run SGD optimizer with a momentum of $0.9$ on both datasets with polynomial learning rate decay.
All experimental results are averaged from 3 runs on 3 different labeled / unlabeled splits and obtained from a single RTX 3090.




\textbf{Evaluation protocol.} 
Following common practice, we feed one single image of its original size to the network for testing, and report mean Intersection-over-Union (mIoU). 
In addition, 
we also report the relative improvements of each method over its reported supervised-only baselines. 
We does not include  \cite{chen2021semi,hu2021semi,wang2022semi} in our comparison because they use sliding window inference and ensemble multiple results for testing (which largely increases mIoU \cite{liu2021perturbed}). 

\subsection{Comparison with state of the art}
\label{secsec:main_exp}
We summarize the results in 
in Table~\ref{tab:city} and Table~\ref{tab:voc} and make comparisons in two aspects. 

\textbf{Improvement over supervised-only baseline.} 
Compared with the baseline where only labeled training data are used, MVCC consistently produces higher mIoU across all settings. 
Specifically, under 1/32, 1/16, 1/8, and 1/4 labeled training data, we witness +16.7\%, +13.2\%, +8.9, and +4.8\% improvement on Cityscapes, and +11.8\%, 9.4\%, +6.2\%, and +4.6\% improvement on Pascal VOC. 
Compared to existing methods, the improvement itself is very competitive, especially under lower labeled data ratios (1/32 and 1/16 splits), which are challenging but arguably more similar to the real-world scenario (with abundant unlabeled data). ST++ \cite{yang2021st++} achieves more improvement under the 1/4 setting for both datasets. ELN \cite{kwon2022semi} reports improvement significantly higher than other methods under 1/8 and 1/4 setting on Cityscapes, but its improvement is lower on Pascal VOC.

\begin{table}[]
\caption{Comparison with state of the arts on Pascal VOC. All notations are the same with Table~\ref{tab:city}. 
}
\label{tab:voc}
\centering
\setlength{\tabcolsep}{3pt}
\small
\begin{tabular}{l|cc|cc|cc|cc|c}
\toprule
Method            & \multicolumn{2}{c|}{1/32} & \multicolumn{2}{c|}{1/16} & \multicolumn{2}{c|}{1/8} & \multicolumn{2}{c|}{1/4} & Oracle \\ \hline
ECS \cite{mendel2020semi}        & -           & -          & -           & -          & 70.2       & +5.0        & 72.6        & +2.8       & 76.3   \\
Alonso \etal \cite{alonso2021semi} & -           & -          & -           & -          & 71.8       & -          & -           & -         & 75.9   \\
ELN \cite{kwon2022semi}        & -           & -          & -           & -          & 73.2       & +5.6       & 74.6       & +4.1      & 76.6   \\
ST++ \cite{yang2021st++}       & -           & -          & 72.6        & +7.8        & 74.4       & +6.1        & 75.4        & \textbf{+4.9}       & -      \\
DBSN$^*$ \cite{yuan2021simple}       & 64.6        & +5.4        & 69.8        & +5.9        & -          & -          & -           & -         & \textbf{76.8}   \\
CAC \cite{lai2021semi}        & 65.1        & +5.9        & 70.1        & +6.2        & 72.4       & +4.1        & 74.0        & +2.8       & 76.3   \\
CPS$^*$ \cite{chen2021semi}        & 64.8        & +5.6        & 68.2        & +4.3          & -       & -          & -       & -         & \textbf{76.8}      \\
USRN \cite{guan2022unbiased}       & 68.6        & +9.4        & 72.3        & +8.4        & -          & -          & -           & -         & \textbf{76.8}   \\ \hline
MVCC (Ours)       & \textbf{70.9} \footnotesize{$\pm$ 0.6}       & \textbf{+11.8}       & \textbf{73.8} \footnotesize{$\pm$ 0.3}       & \textbf{+9.4}        & \textbf{75.3} \footnotesize{$\pm$ 0.3}      & \textbf{+6.2}        & \textbf{75.8} \footnotesize{$\pm$ 0.2}       & +4.6       & 75.9 \footnotesize{$\pm$ 0.2} \\
\bottomrule
\end{tabular}
\end{table}

\textbf{Overall system performance.}
With a reasonably effective fully supervised oracle, 
on both datasets, MVCC learning achieves very competitive accuracy. 
On Cityscapes, we report 72.0\%, 74.8\%, 76.8\%, and 77.2\% mIoU under 1/32, 1/16, 1/8, and 1/4 labeled training data, respectively. These results compare favorably against results of existing methods under similar settings. 
On Pascal VOC, MVCC produces 70.9\%, 73.8\%, 75.3\%, and 75.8\% mIoU, respectively, which are higher than methods with similar network architectures.
It is noteworthy that the proposed method works particularly well compared to competing methods under low labeled data ratios. 
On the other hand, for higher labeled data ratios, 
with 1/8 labeled training data, MVCC produces an overall result just 0.6\% lower than the fully supervised oracle on both datasets, further demonstrating its effectiveness. 

\begin{figure}
\centering
\begin{subfigure}[b]{0.24\linewidth}
\includegraphics[width=\linewidth]{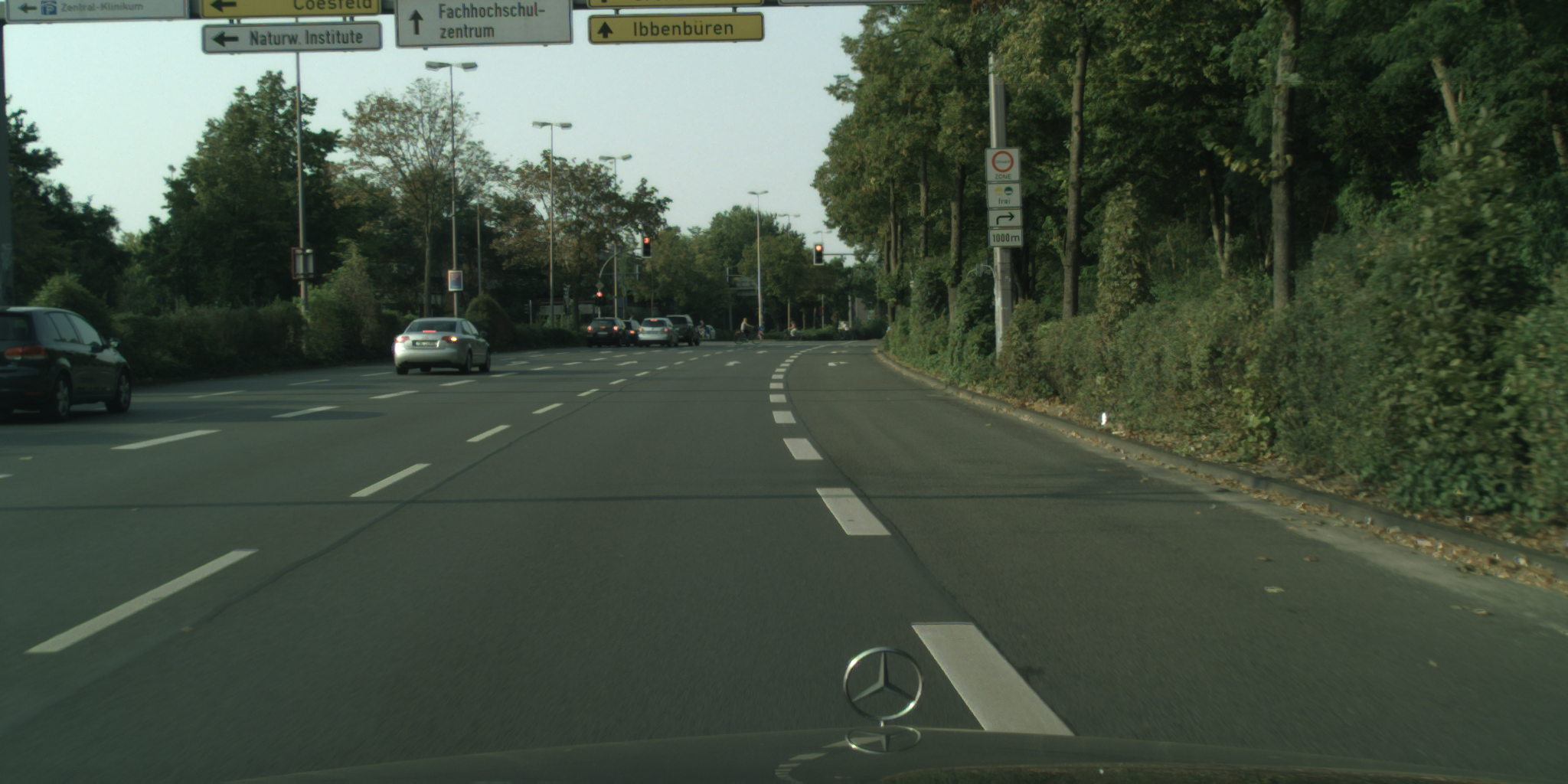}
\includegraphics[width=\linewidth]{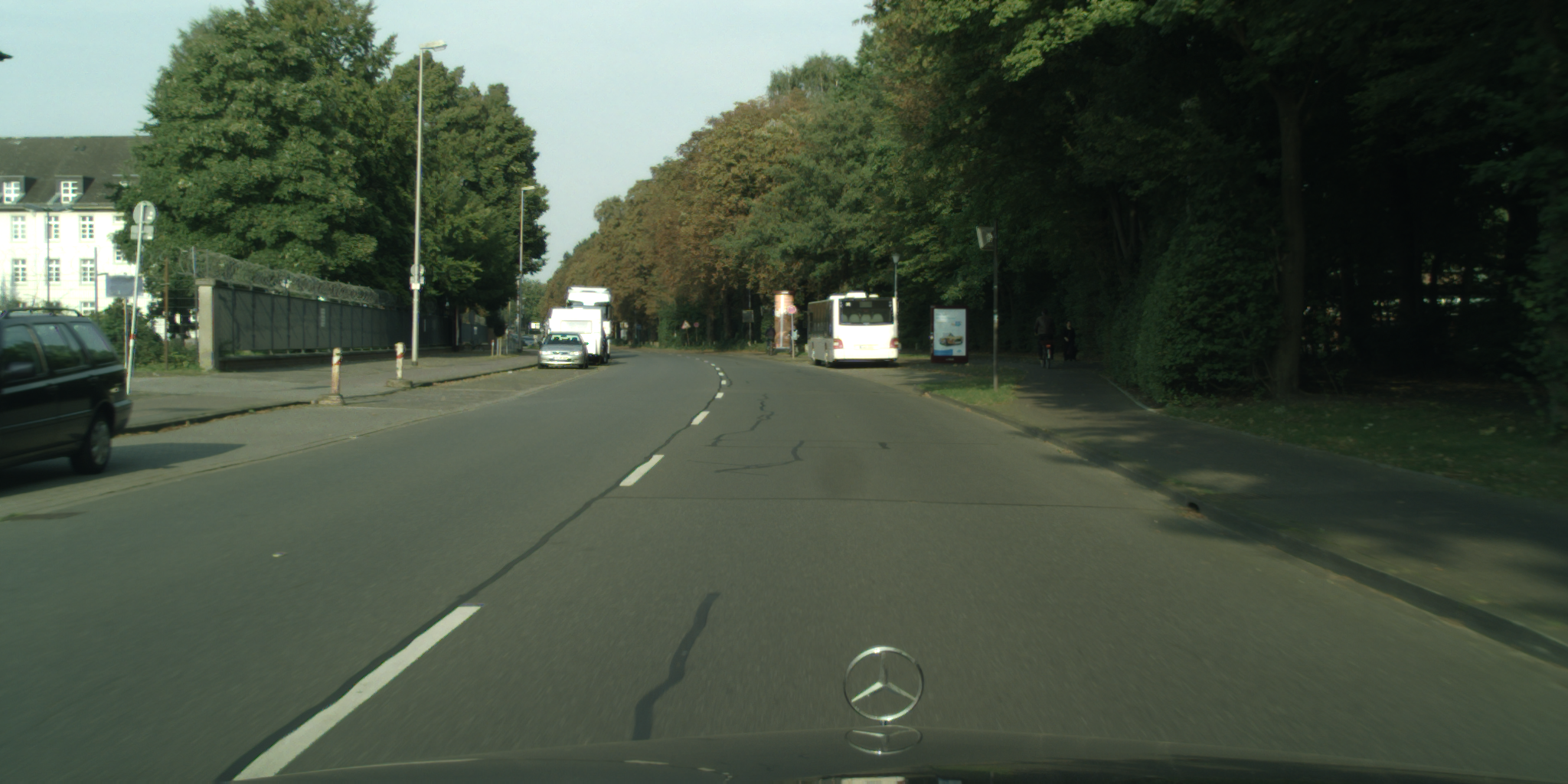}
\includegraphics[width=\linewidth]{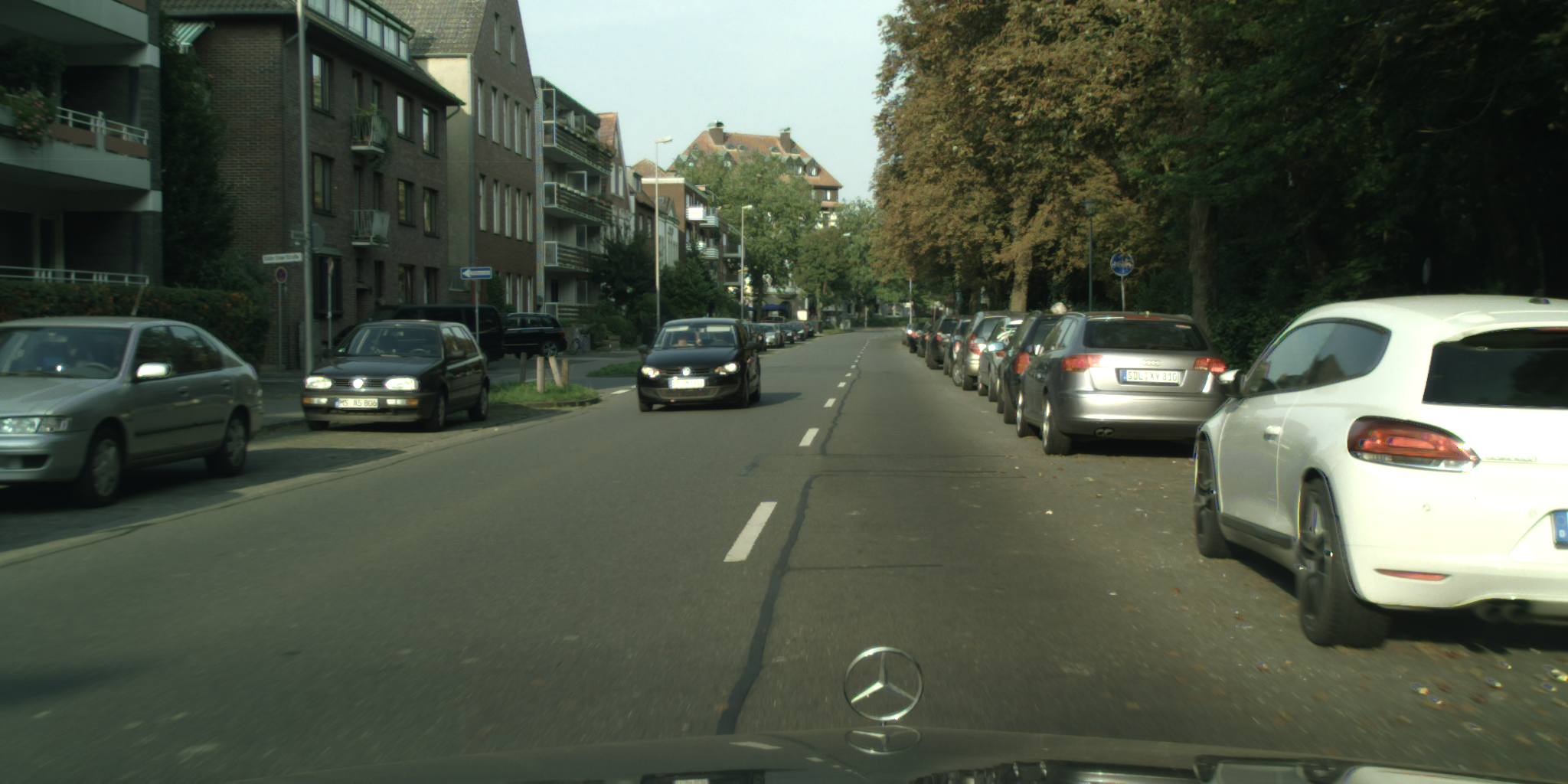}
\end{subfigure}
\begin{subfigure}[b]{0.24\linewidth}
\includegraphics[width=\linewidth]{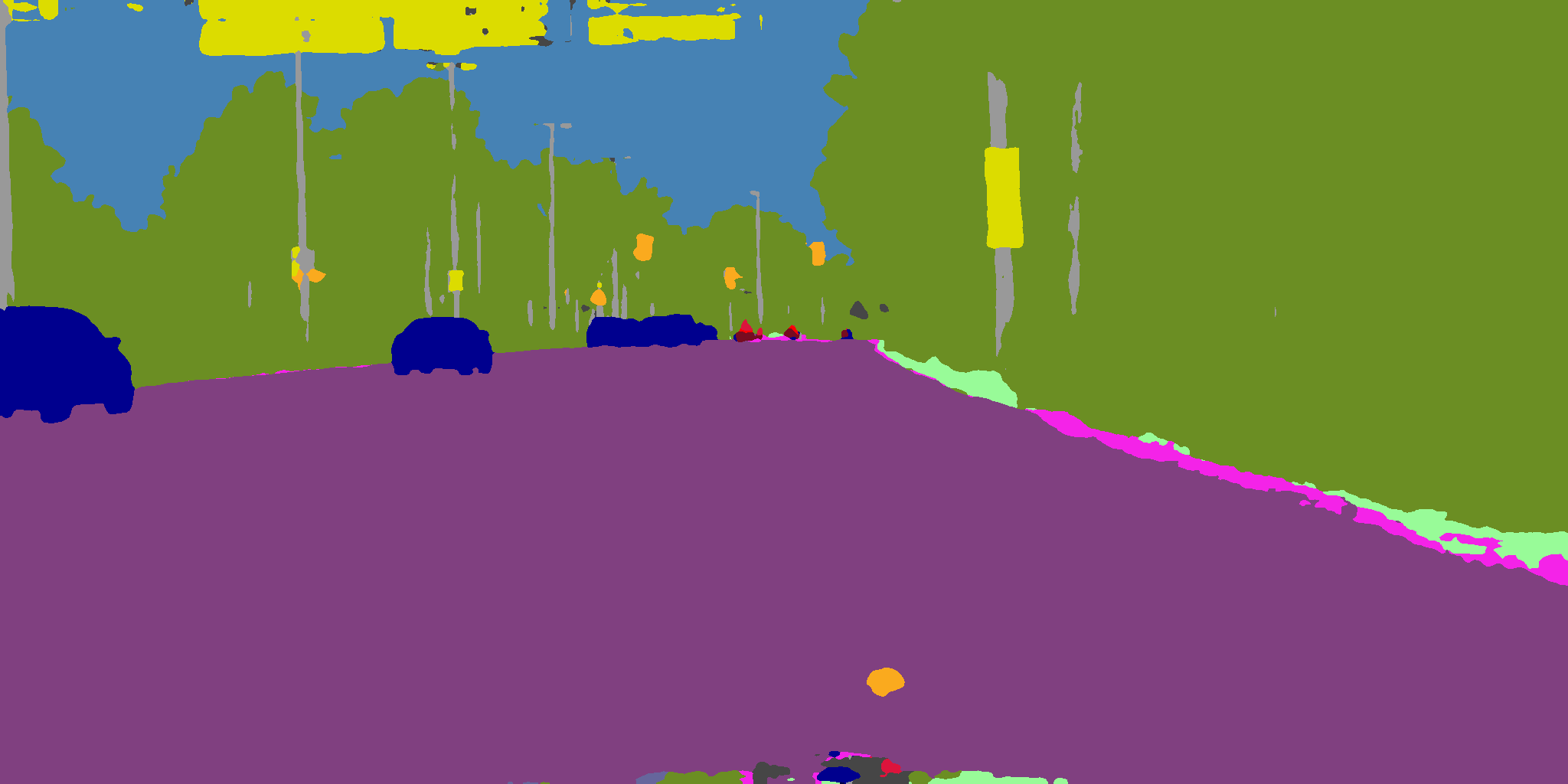}
\includegraphics[width=\linewidth]{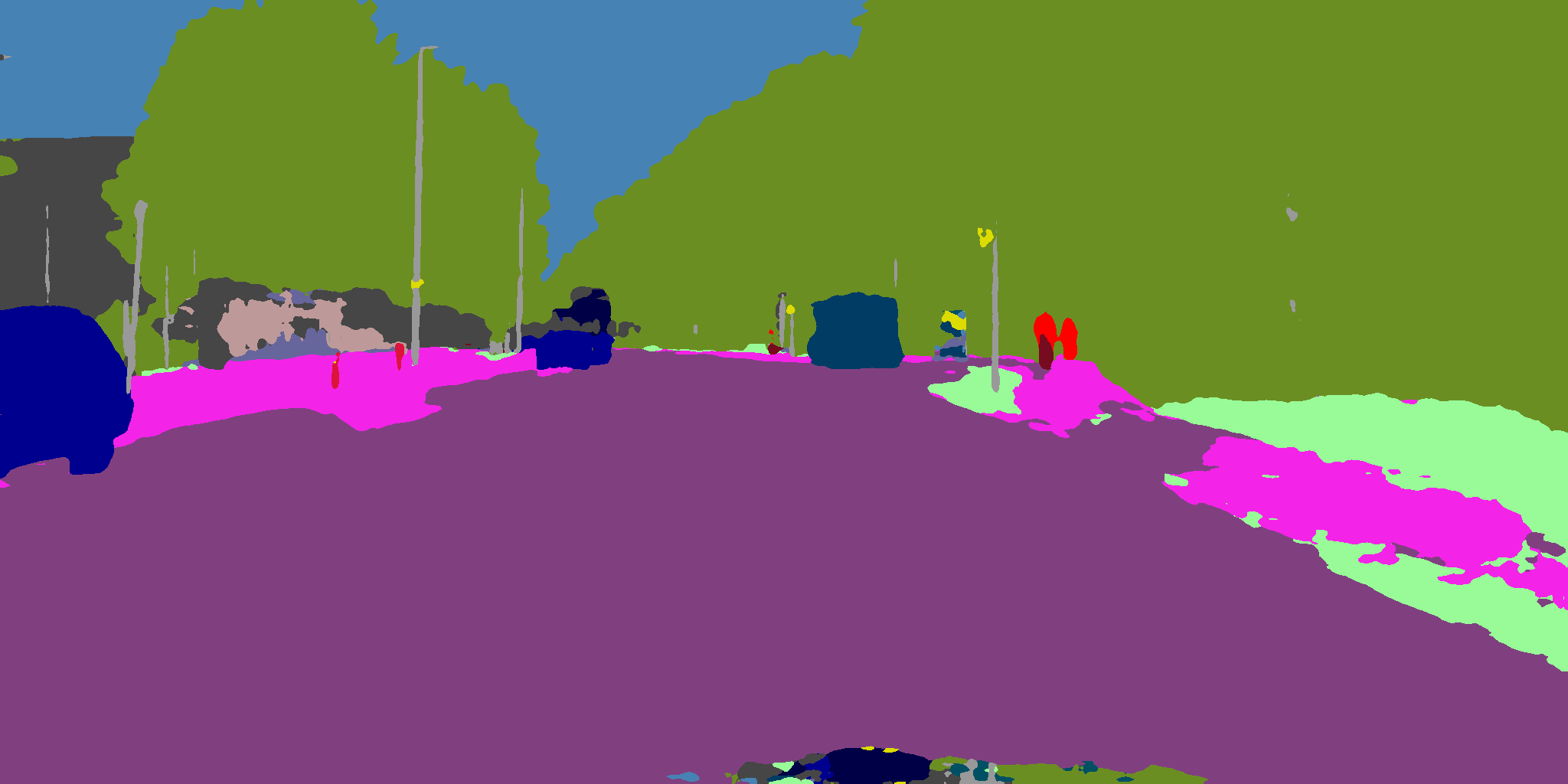}
\includegraphics[width=\linewidth]{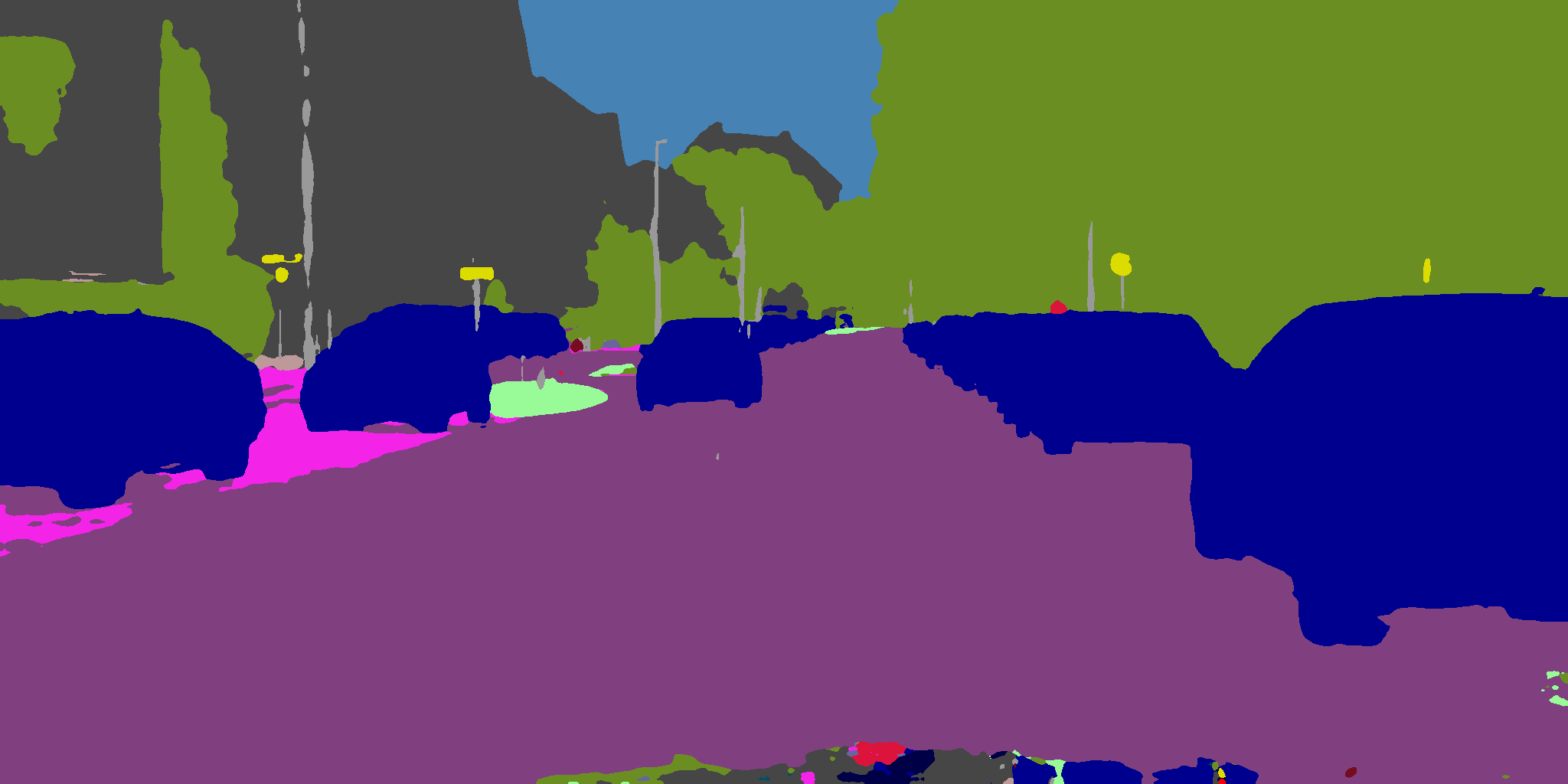}
\end{subfigure}
\begin{subfigure}[b]{0.24\linewidth}
\includegraphics[width=\linewidth]{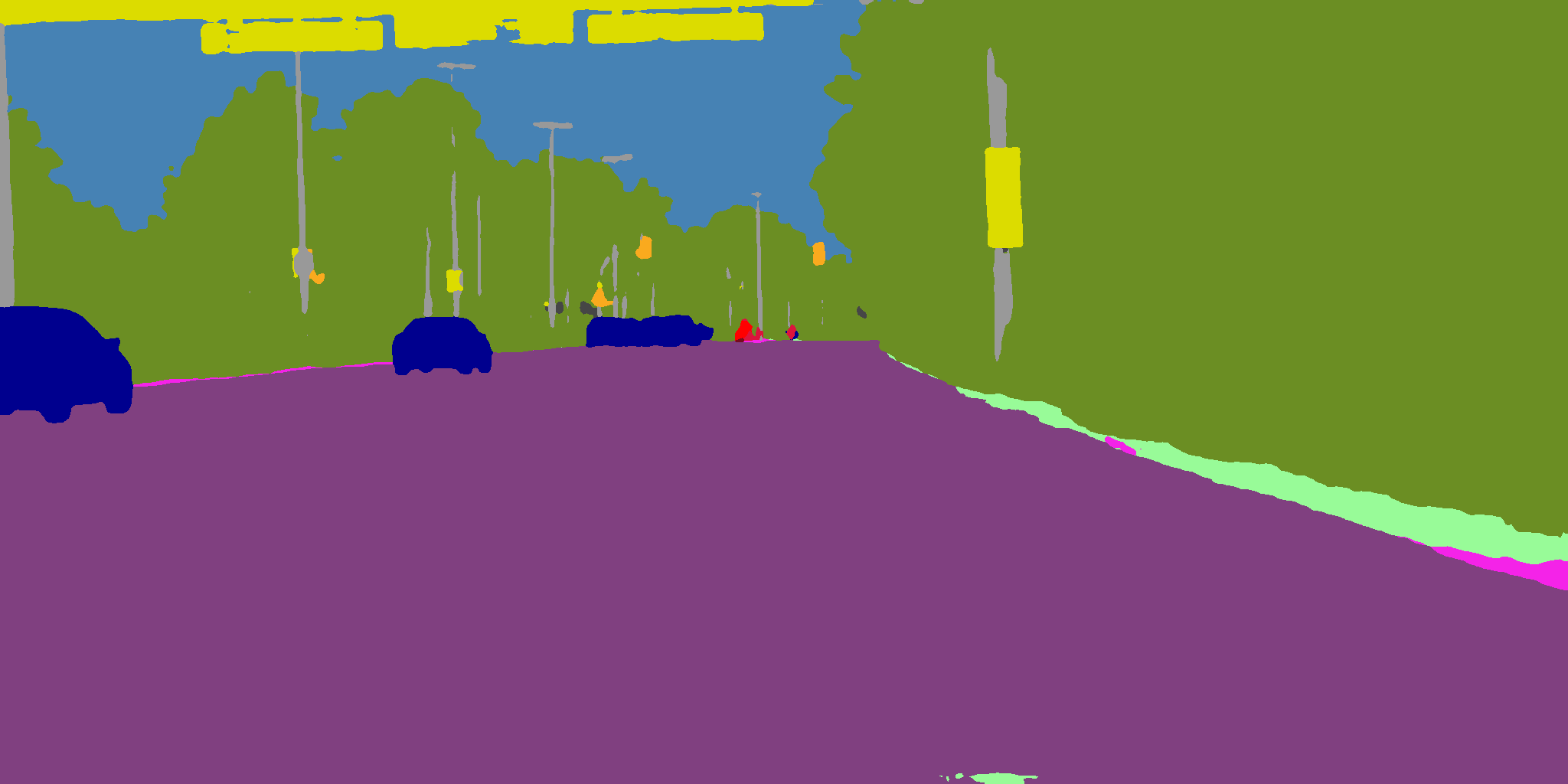}
\includegraphics[width=\linewidth]{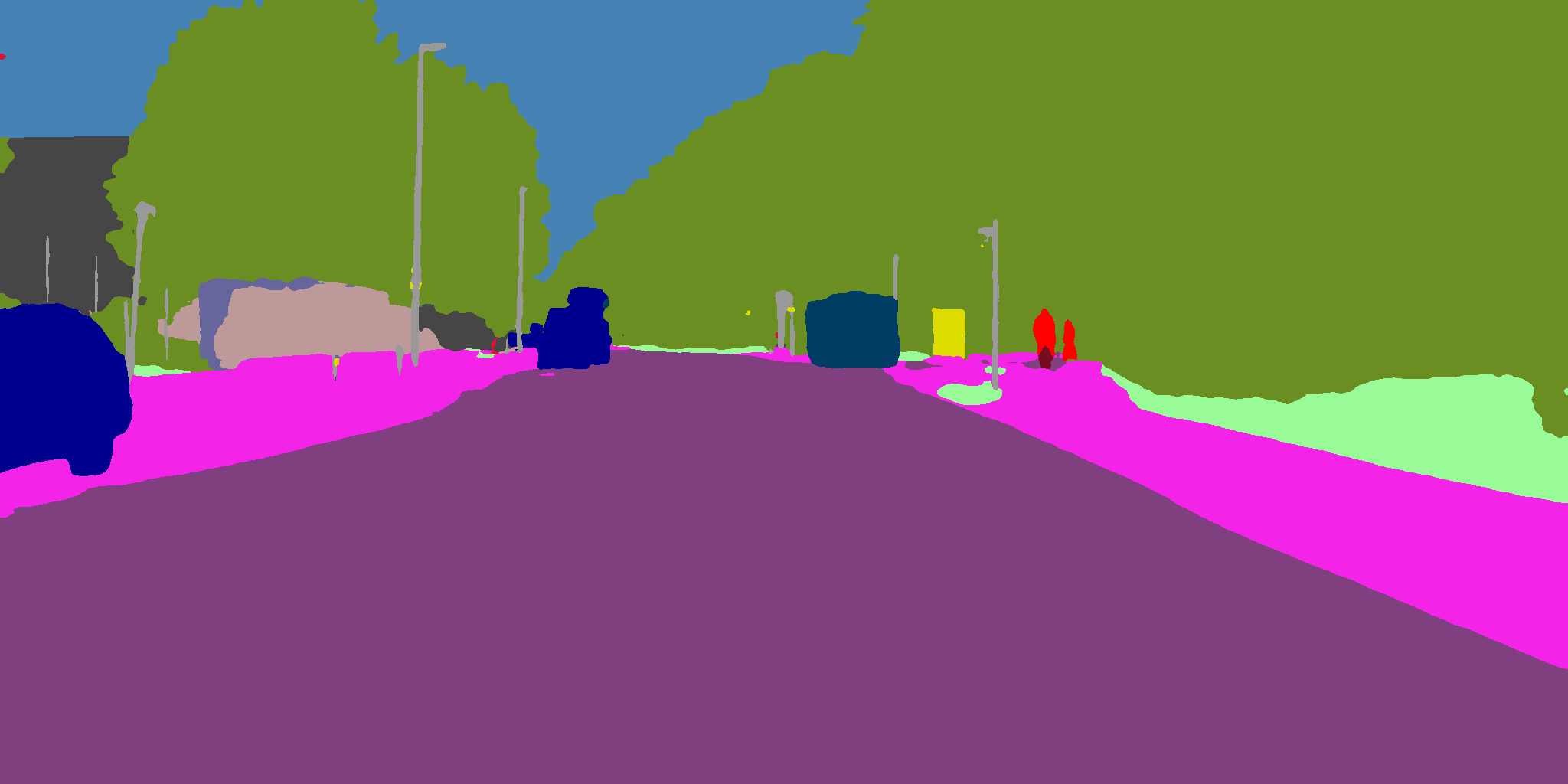}
\includegraphics[width=\linewidth]{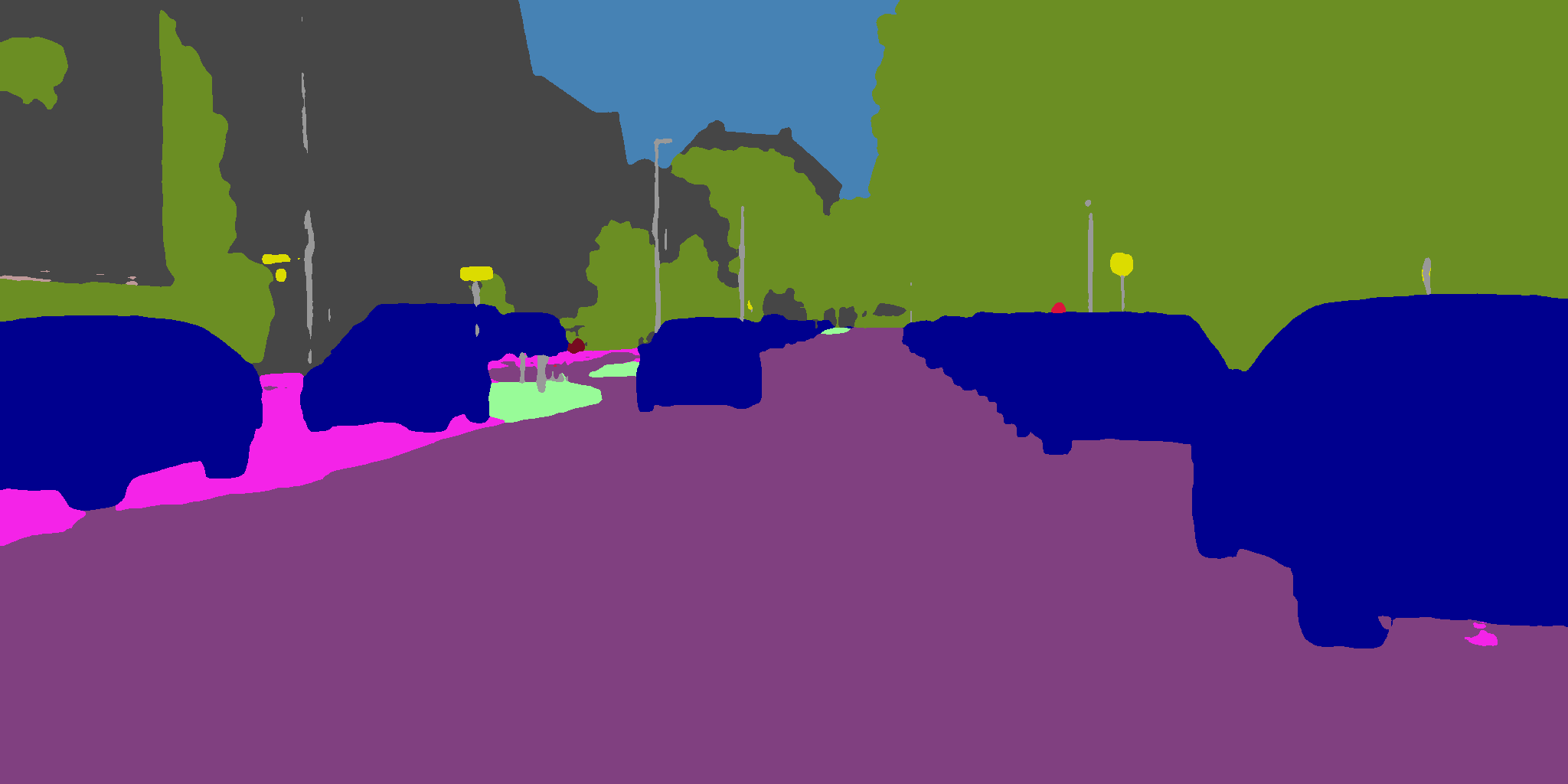}
\end{subfigure}
\begin{subfigure}[b]{0.24\linewidth}
\includegraphics[width=\linewidth]{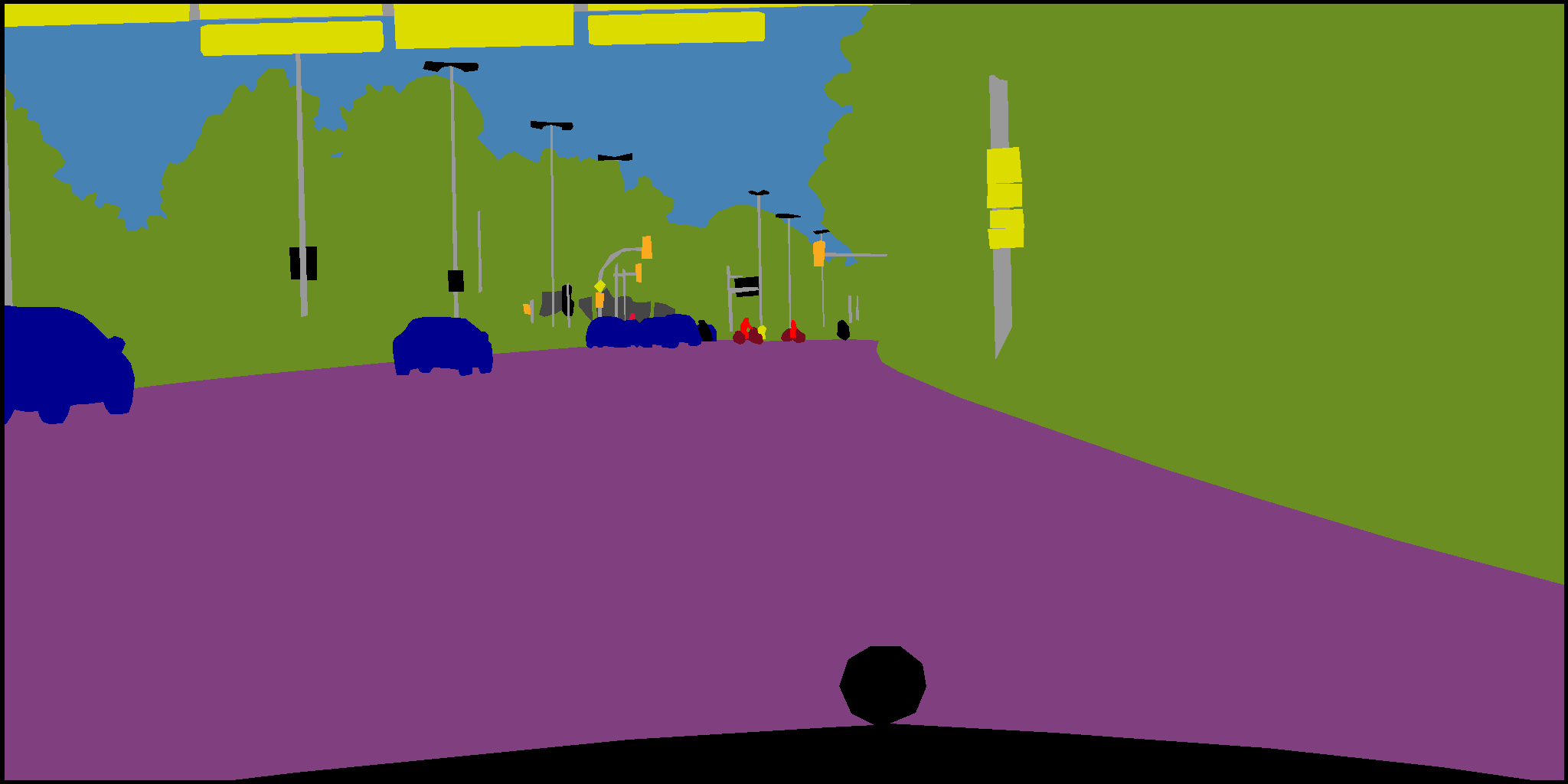}
\includegraphics[width=\linewidth]{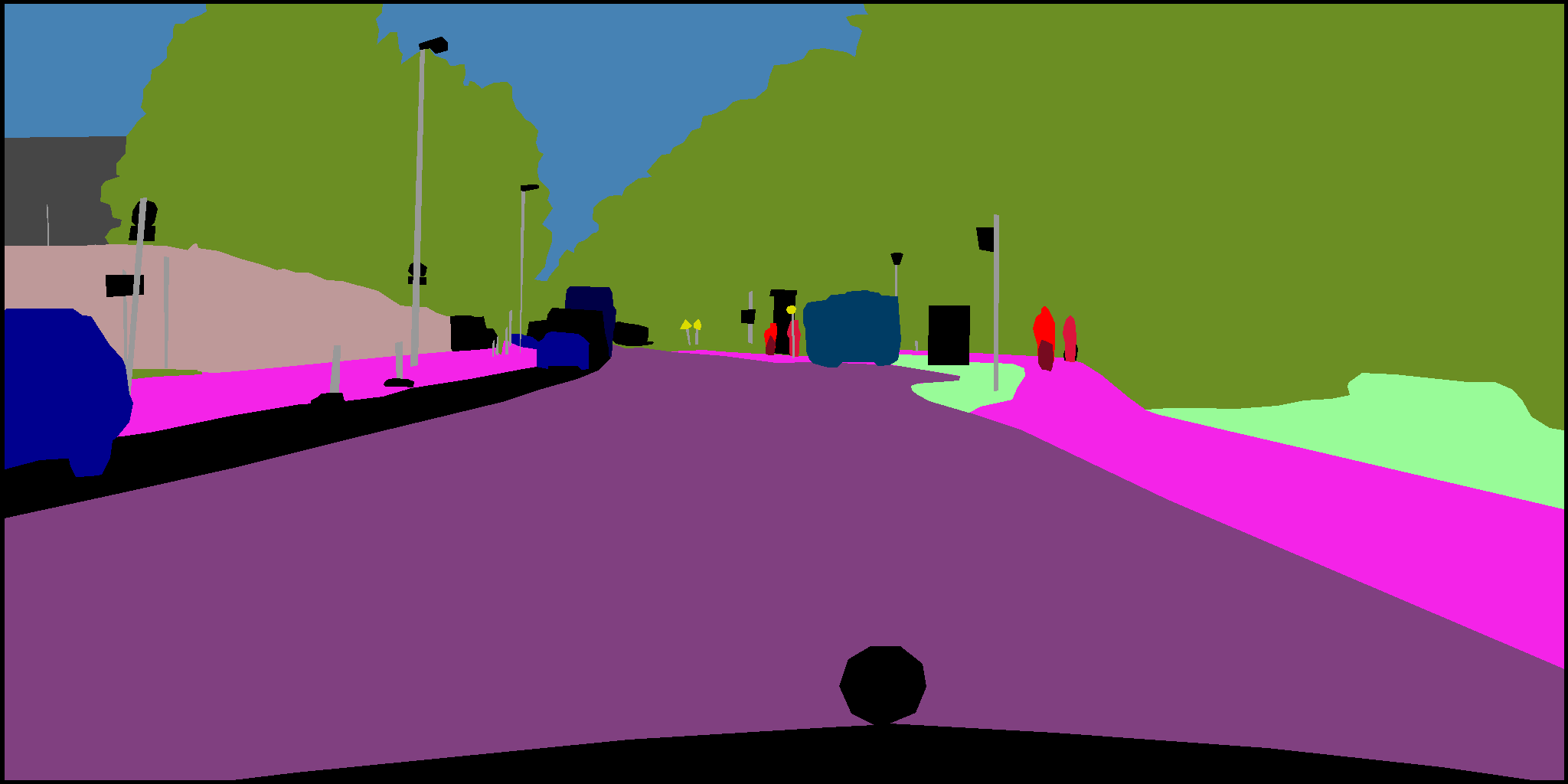}
\includegraphics[width=\linewidth]{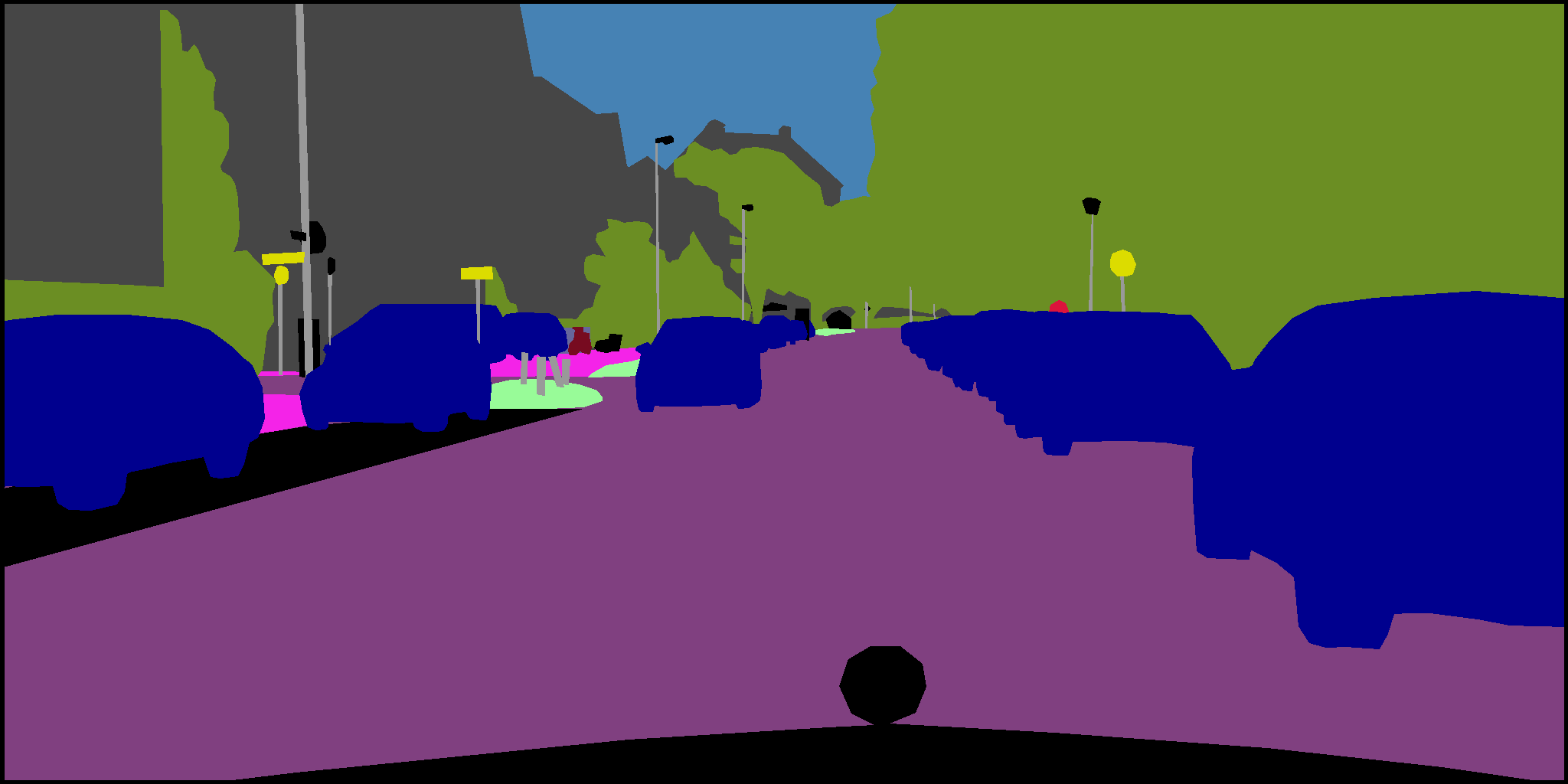}
\end{subfigure}
\caption{Result comparison on Cityscapes with  1/32 labeled data. From left to right: input images, supervised-only baseline, MVCC learning, and ground truth. }
\label{fig:results_demo}
\end{figure}

We show visual result demos in Fig.~\ref{fig:results_demo}. Compared to the supervised-only baseline, in MVCC, pixels of the same object (\eg, trees, pavement, wall) are given more consistent predictions.

\begin{figure}
\begin{subfigure}[b]{0.32\linewidth}
\includegraphics[width=\linewidth]{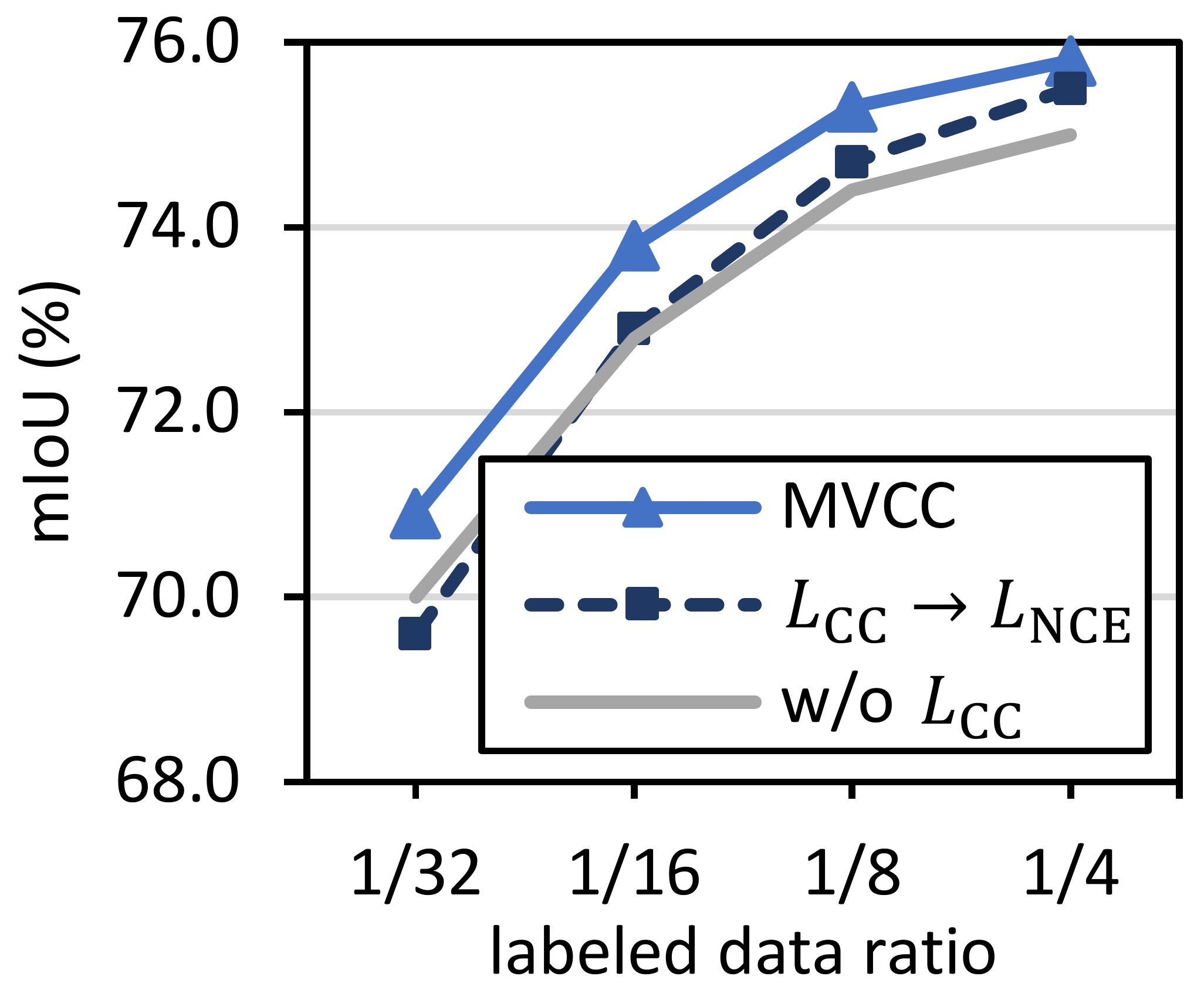}
\end{subfigure}
\hfill
\begin{subfigure}[b]{0.32\linewidth}
\includegraphics[width=\linewidth]{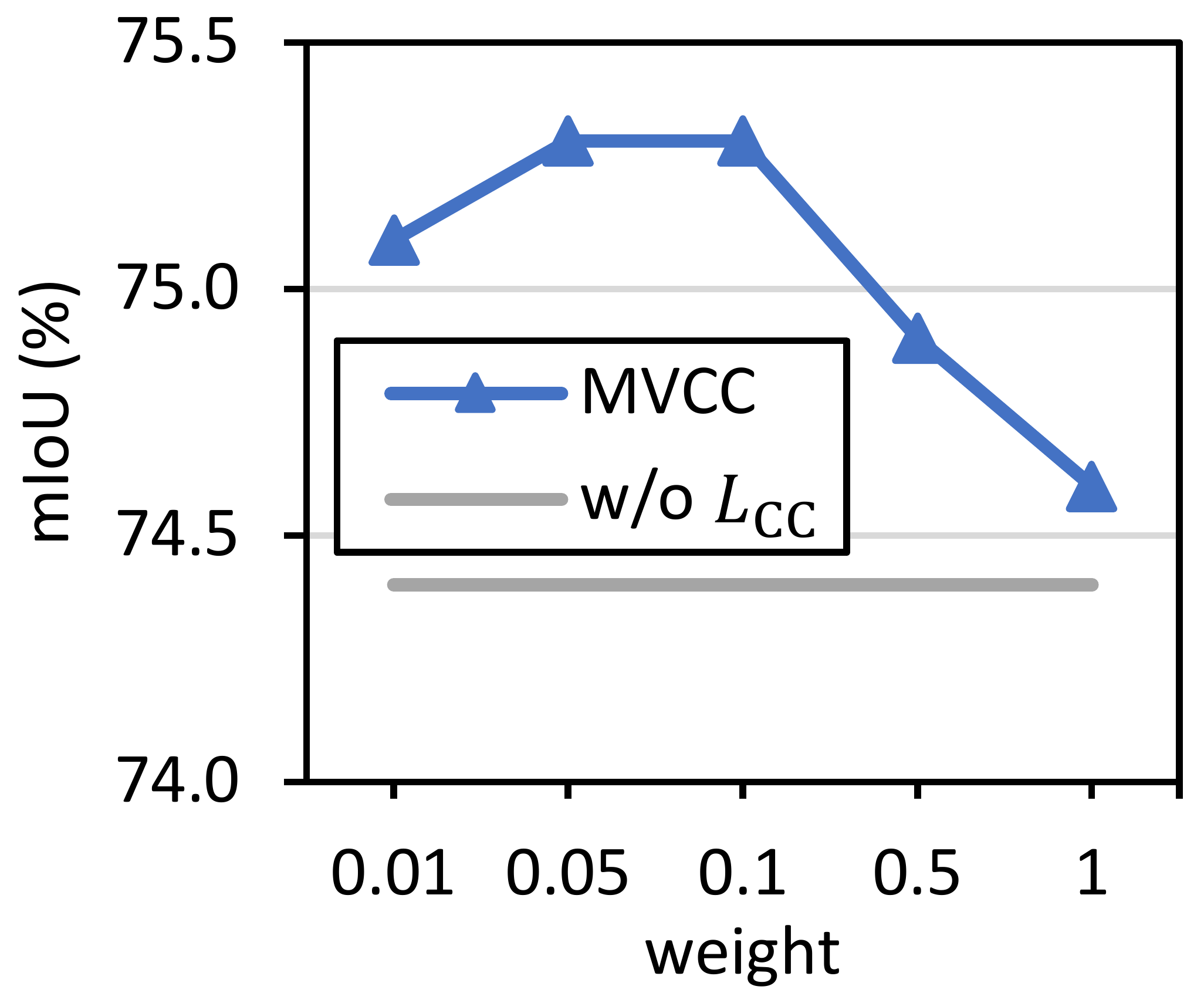}
\end{subfigure}
\hfill
\begin{subfigure}[b]{0.32\linewidth}
\includegraphics[width=\linewidth]{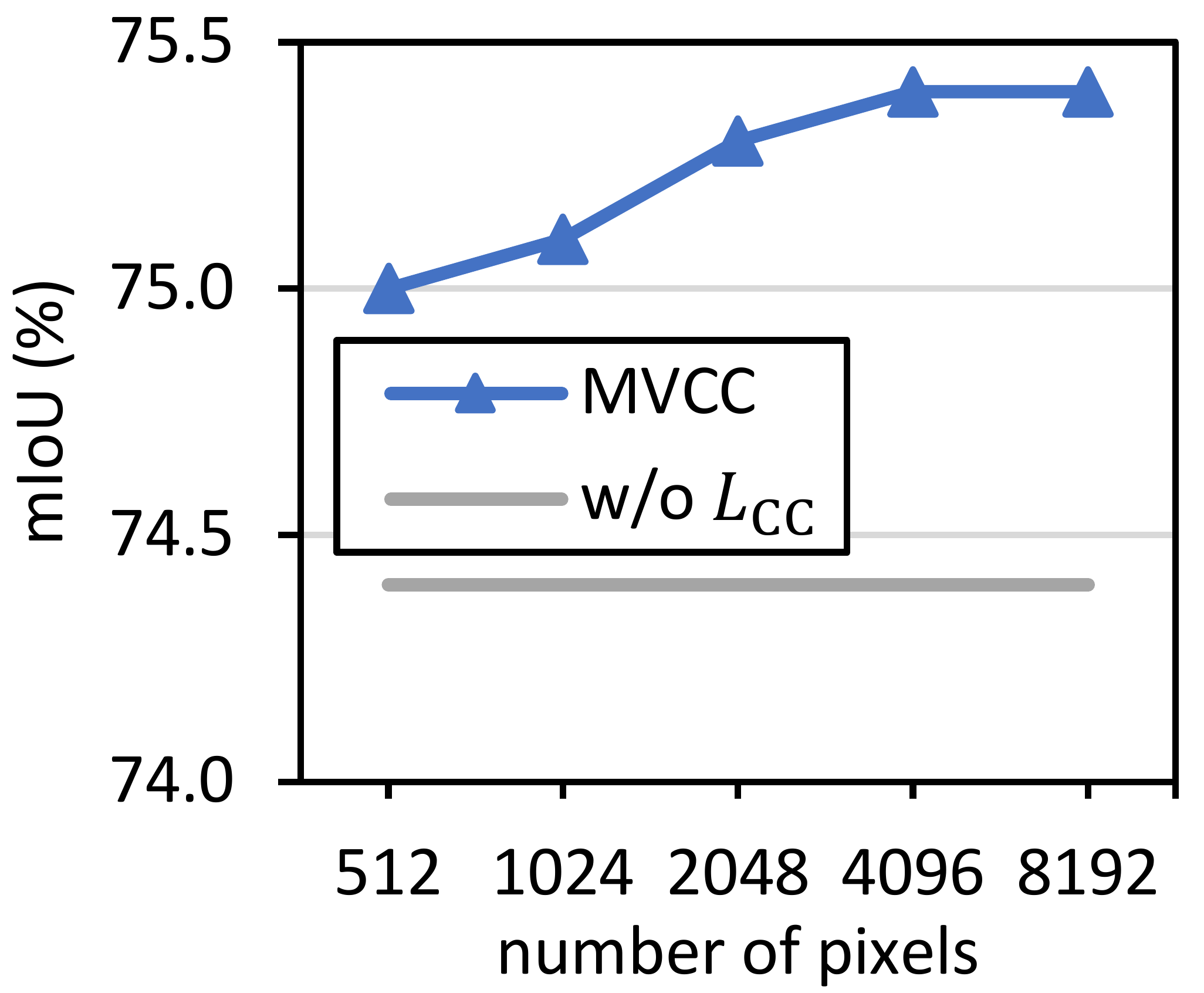}
\end{subfigure}
\caption{Variant and hyperparameter analysis on Pascal VOC. 
\textbf{Left}: Comparison between the proposed method (``MVCC''), its contrastive learning variant (``$\mathcal{L}_\text{CC}\rightarrow\mathcal{L}_\text{NCE}$''), and its consistency learning only variant (``w/o $\mathcal{L}_\text{CC}$''). 
\textbf{Middle}: Impact of the weight for the correlation consistency loss (Eq.~\ref{eq:loss_overall}).
\textbf{Right}: Impact of the number of sampled pixels $N$ (Section~\ref{secsecsec:sampling}). We use 1/8 labeled data for the middle and right figures. 
}
\label{fig:combined_variant}
\end{figure}

\begin{wraptable}{R}{0.42\linewidth}
\vspace{-4mm}
\caption{Comparison with variants on Pascal VOC (1/8 labeled data ratio). }
\label{tab:variant}
\centering
\setlength{\tabcolsep}{3pt} 
\small
\begin{tabular}{lc}
\toprule
Method                                & mIoU (\%)           \\ \hline
supervised-only baseline               & 69.1                \\
MVCC (full system)              & 75.3                \\
w/o $\mathcal{L}_\text{CC}$                 & 74.4     \\
$\mathcal{L}_\text{CC}\rightarrow\mathcal{L}_\text{NCE}$   & 74.7 \\
same geometric augmentation                 & 74.8                \\
w/o view-coherent CutMix                          & 55.1                \\
w/ projection head                    & 75.4                \\
w/o label smoothing                    & 75.2                \\
\bottomrule
\end{tabular}
\end{wraptable}

\subsection{Variant and ablation study}
\label{secsec:ablation}

We examine other design choices by comparing them with the proposed method in Table~\ref{tab:variant} and Fig.~\ref{fig:combined_variant}. 
If not specified, we use Pascal VOC with 1/8 labeled data. 

\textbf{Effectiveness of correlation consistency loss $\mathcal{L}_\text{CC}$.}
Without $\mathcal{L}_\text{CC}$ (``w/o $\mathcal{L}_\text{CC}$''), our method reduces to its consistency learning variant. 
In Fig.~\ref{fig:combined_variant}, this variant has consistently lower mIoU under every labeled data ratio. Specifically, under 1/8 labeled data, mIoU drops by 0.9\% (Table~\ref{tab:variant}).  
Such results verify the effectiveness of $\mathcal{L}_\text{CC}$.


\textbf{Comparison between correlation consistency $\mathcal{L}_\text{CC}$ 
and InfoNCE loss  $\mathcal{L}_\text{NCE}$.}
When we replace $\mathcal{L}_\text{CC}$ with $\mathcal{L}_\text{NCE}$ (``$\mathcal{L}_\text{CC}\rightarrow\mathcal{L}_\text{NCE}$''), our method becomes its contrastive learning variant. This variant follows the implementation in \cite{wang2021exploring}, and all other settings are the same as the proposed MVCC. We have two observations. First, this variant is consistently inferior to our method under all the ratios. Specifically, under 1/8 ratio, the mIoU difference is 0.6\%. Second, 
under relatively high labeled data ratios, \eg, 1/8, we find this variant more effective than the consistency only variant ``w/o $\mathcal{L}_\text{CC}$'', demonstrated by a 0.3\% higher mIoU. 
Third, under lower labeled data ratios, its effectiveness is limited, which is consistent with the analysis in Section~\ref{secsec:discussion}. 



\textbf{Effectiveness of {image-level} augmentation.}
As shown in Fig.~\ref{fig:system}, the proposed MVCC approach use different {image-level} geometric augmentation for different views. In Table~\ref{tab:variant}, we examine its benefit with the ``same geometric augmentation'' variant, where  two views of an unlabeled image are under the same geometric augmentation. 
Observing a $0.5\%$ mIoU drop, we verify it is useful for different views to undergo different image-level augmentation. 

\textbf{Effectiveness of view-coherent {region-level} augmentation on different geometrically augmented views.}
For different geometrically augmented views, randomly applying {region-level} augmentation breaks area correspondence between views (Section~\ref{secsecsec:region-level}).
In our experiment, when randomly applying CutMix to the two views (``w/o view-coherent CutMix'' in Table~\ref{tab:variant}),  segmentation accuracy drops significantly ($20.2\%$ in mIoU). 


\textbf{Influence of an additional projection head.}
As indicated by the ``w/ projection head'' variant in Table~\ref{tab:variant}, further including a projection head and restricting correlation consistency loss on the projected features does not bring significant improvement. 

\textbf{Influence of label smoothing.}
Removing label smoothing from our system (``w/o label smoothing'') does not cause a major performance change. With that said, experiments find that label smoothing helps to stabilize training: it reduces the standard deviation from $\pm$0.8\% to $\pm$0.3\% mIoU.

\textbf{Hyperparameter analysis.} 
In Fig.~\ref{fig:combined_variant}, we assess the system sensitivity to the weight of $\mathcal{L}_\text{CC}$ (Eq.~\ref{eq:loss_overall}) and the number of pixels $N$ (Section~\ref{secsecsec:sampling}). 
For the former, its optimal value is between 0.05 and 0.1, and decreasing or increasing it leads to mIoU drops. We set this weight as 0.1 in our experiments to align with previous contrastive methods \cite{wang2021exploring,zhou2021c3,alonso2021semi,hu2021region}. 
For the latter, we observe a slight mIoU increase when its value increases. We choose $N=2048$ as a trade-off between accuracy and computation complexity. 
Note that we adopt the same hyperparameter values across all settings. 

\section{Conclusion}
In this paper, we analyze the \textit{pros} and \textit{cons} of consistency and contrastive learning in the context of semi-supervised semantic segmentation. Our analysis motivates us to look for rich and robust supervision on unlabeled data: using possibly many pixel pairs and forcing their pairwise similarity to be constant across views. We integrate this supervision signal in a multi-view correlation consistency (MVCC) approach, which is shown to outperform its consistency learning only and contrastive learning variants. This approach additionally contains a view-coherent data augmentation strategy for the dense prediction task, which allows for differentiable and invertible \textit{image-level} geometric augmentation and view-coherent \textit{region-level} augmentation. 
In a series of semi-supervised settings, our method has very competitive performance compared with the state of the art. 


{\small
\bibliographystyle{plain}
\bibliography{main}
}

\newpage

\appendix



\end{document}